\documentclass[11pt]{article}

\usepackage[final]{acl}

\usepackage{times}
\usepackage{latexsym}
\usepackage{soul}
\usepackage[T1]{fontenc}

\usepackage[utf8]{inputenc}

\usepackage{microtype}
\usepackage{bm}
\usepackage{inconsolata}

\usepackage{graphicx}
\usepackage{multirow}
\usepackage{placeins}
\usepackage[most]{tcolorbox}

\newcommand{\prompt}[4]{%
  \begin{figure*}[ht]
    \centering
    \vspace{1em}
    \noindent\colorbox{gray!20}{%
      \parbox{0.98\textwidth}{
        \small\textbf{#1:} \\ 
        #2
      }
    }
    \vspace{1em}
    \caption{#3}
    \label{fig:#4}
  \end{figure*}
}

\title{MM-BizRAG: Rethinking Multimodal Retrieval-Augmented Generation for General Purpose Enterprise Q\&A}

\author{
 \textbf{Hanoz Bhathena \textsuperscript{\P}}
 \textbf{Parin Rajesh Jhaveri \textsuperscript{\P}}
 \textbf{Rohan Mittal \textsuperscript{\P}}
 \textbf{Prateek Singh \textsuperscript{\P}}
\\
 \textbf{Aymen Kallala \textsuperscript{\P}}
 \textbf{Rachneet Kaur \textsuperscript{\P}}
 \textbf{Yiqiao Jin \textsuperscript{||}}
 \textbf{Zhen Zeng \textsuperscript{\P}}
\\
 \textbf{Adwait Ratnaparkhi \textsuperscript{\P}}
 \textbf{Denis Kochedykov \textsuperscript{\P}}
\\
\\
 \textsuperscript{\P}JPMorgan Chase \& Co.
 \textsuperscript{||}Georgia Institute of Technology
\\
 \small{
\href{mailto:hanoz.bhathena@jpmchase.com}{hanoz.bhathena@jpmchase.com}
 }
}

\begin{document}
\maketitle

\begin{abstract}
Recent advances in multimodal retrieval-augmented generation (MM-RAG) have shifted toward minimal parsing, relying on page-level images for producing retriever embeddings and for answer generation. While efficient, this trend often neglects explicit handling of the rich, structured information in complex enterprise documents, instead depending on pre-trained embeddings or vision-language models to implicitly capture such structure. In this work, we take a more direct approach: \textbf{MM-BizRAG} proactively extracts and represents document structure via a \textit{document structure-aware split} that dynamically routes documents through orientation-specific ingestion pipelines, applying explicit layout-aware parsing for vertically structured documents (e.g., reports) and holistic page-level representations for horizontally structured documents (e.g., slide decks). A unified LLM-driven artifact transformation pipeline with placeholder-based positional alignment preserves natural reading order, while inference-time multimodal assembly decouples retrieval representations from generation context, enabling richer, more grounded answers without any finetuning requirement. Through experiments on a large, heterogeneous enterprise dataset and two public benchmarks (\textsc{SlideVQA} and \textsc{FinRAGBench-V}), MM-BizRAG consistently outperforms state-of-the-art vision-centric baselines by up to 32\% points, with especially strong gains on report-style layouts. Furthermore, we introduce \textbf{FastRAGEval}, a single-call LLM Judge metric for fine-grained generative recall that halves RAGChecker's cost while achieving stronger human alignment.
\end{abstract}

\section{Introduction}

Modern RAG systems have evolved beyond text-only inputs to incorporate multiple modalities including images, videos, and complex document graphs~\cite{abootorabi2025ask, mei2025survey, gao2025scaling, edge2024local}, enabling retrieval and reasoning across diverse data types. This evolution has been driven by significant advances across various RAG pipeline components, for example improved document parsing and layout analysis enabled by pre-trained document layout models, long context~\cite{nussbaum2024nomic} and multimodal embedding models~\cite{ma2024unifyingmultimodalretrievaldocument, xu2025survey, jiang2024vlm2vec, yu2024visrag, gunther2025jina}, and multimodal LLMs capable of processing interleaved text and images to generate responses~\cite{han2025multimodal}.

Enterprise documents span various file types including PDFs, DOCX, PPTX, HTML pages, each containing combinations of text, tables, and images, often interspaced within complex layouts. Recent MM-RAG approaches have shifted toward minimal parsing, relying solely on page images for both embeddings for retrieval and as inputs to VLMs for answer generation. While efficient and appealing in their simplicity, eliminating the need for complex layout parsing by offloading structural understanding to VLMs and multimodal embeddings, these approaches empirically fall short, as pre-trained models struggle to implicitly capture the rich, structured information embedded in complex enterprise documents typically not part of their training data. Yet these pipelines often lack standardization~\cite{gao2025scaling, Zhang_2025} in document representation~\cite{sarthi2024raptorrecursiveabstractiveprocessing, jin2025hierarchicaldocumentrefinementlongcontext, yu2024visrag} and ingestion design~\cite{xiong2024benchmarkingretrievalaugmentedgenerationmedicine, asai2023selfraglearningretrievegenerate}.


To overcome this challenge we propose MM-BizRAG, a multimodal RAG framework built around five contributions that, in their cohesive integration, form a materially novel system. At its core, a \textit{document structure-aware split} dynamically routes documents through orientation-specific ingestion pipelines based on vertical (report-style) or horizontal (slide-style) structure; a design unexplored in MM-RAG literature to our knowledge. A unified LLM-driven artifact transformation pipeline with placeholder-based positional alignment for tables and pictures maintains natural reading order throughout downstream inference. Critically, MM-BizRAG decouples artifact representations used during retrieval from those used during answer generation via \textit{inference-time multimodal assembly}, enabling richer context construction without redundant indexing. The cohesive integration of all these components into a single system that works with out-of-box LLMs and encoder models, requiring no finetuning, outperforms vision-centric baselines by up to 32\% points.

To test the efficacy of our pipeline we study the effects of variations in document representation and embedding generation strategy in our ingestion pipeline on downstream performance of the MM-BizRAG framework through a \textit{controlled variant study} that isolates ingestion choices with a fixed inference stack.

The main contributions of our study are:
\begin{itemize}
    \item We introduce \textbf{MM-BizRAG}, a document structure-aware MM-RAG system for heterogeneous, multi-domain enterprise documents, built around three core design innovations: (i) a \textit{document structure-aware split} that dynamically routes documents through orientation-specific ingestion pipelines: layout-aware parsing for vertically structured documents (e.g., reports, filings) and holistic slide-level representations for horizontally structured documents (e.g., slide decks), preserving cross-modal alignment among text, tables, and images; (ii) a unified LLM-driven artifact transformation pipeline with placeholder-based positional alignment (for vertical documents) that maintains natural reading order throughout downstream inference; and (iii) \textit{inference-time multimodal assembly}, which explicitly decouples artifact representations used during retrieval from those assembled for multimodal generation, enabling richer, more contextually grounded generation context to be constructed at inference time without inflating the retrieval index; a core architectural distinction from prior vision-centric MM-RAG approaches.

    \item We present a \textit{controlled variant study} comprising three MM-BizRAG design variants that systematically vary ingestion transformations and embedding strategies while holding the inference pipeline fixed, providing a methodologically novel framework for isolating the impact of ingestion and retrieval representation choices in MM-RAG ablation. We benchmark all variants on a large internal enterprise dataset encompassing diverse file types (PDF, Docx, HTML, PPT), complex layouts, and 30 business domains, as well as two public benchmarks, \textsc{SlideVQA} and \textsc{FinRAGBench-V}, where all variants outperform strong vision-centric MM-RAG baselines (ColPali, VisRAG) by up to 32\% points, with consistent gains across financial, legal, and technical documents in both report and presentation-style formats. Our recommended production variant, TCTE, achieves recall within 1--3\% points of the best configuration at roughly half the latency (for vertical documents).

    \item We propose \textbf{FastRAGEval (FRE)}, a single-call LLM Judge metric for fine-grained generative recall. Unlike RAGChecker (RC)~\cite{ru2024ragcheckerfinegrainedframeworkdiagnosing}, which decomposes answers into atomic claims across two sequential LLM calls, FRE computes precision, recall, and F1 in a single pass; halving cost and latency while achieving stronger alignment with human judgements across multiple correlation measures, validated on 200 human-annotated instances.
\end{itemize}

\begin{figure*}[!h]
\centering
\includegraphics[width=0.9\textwidth]{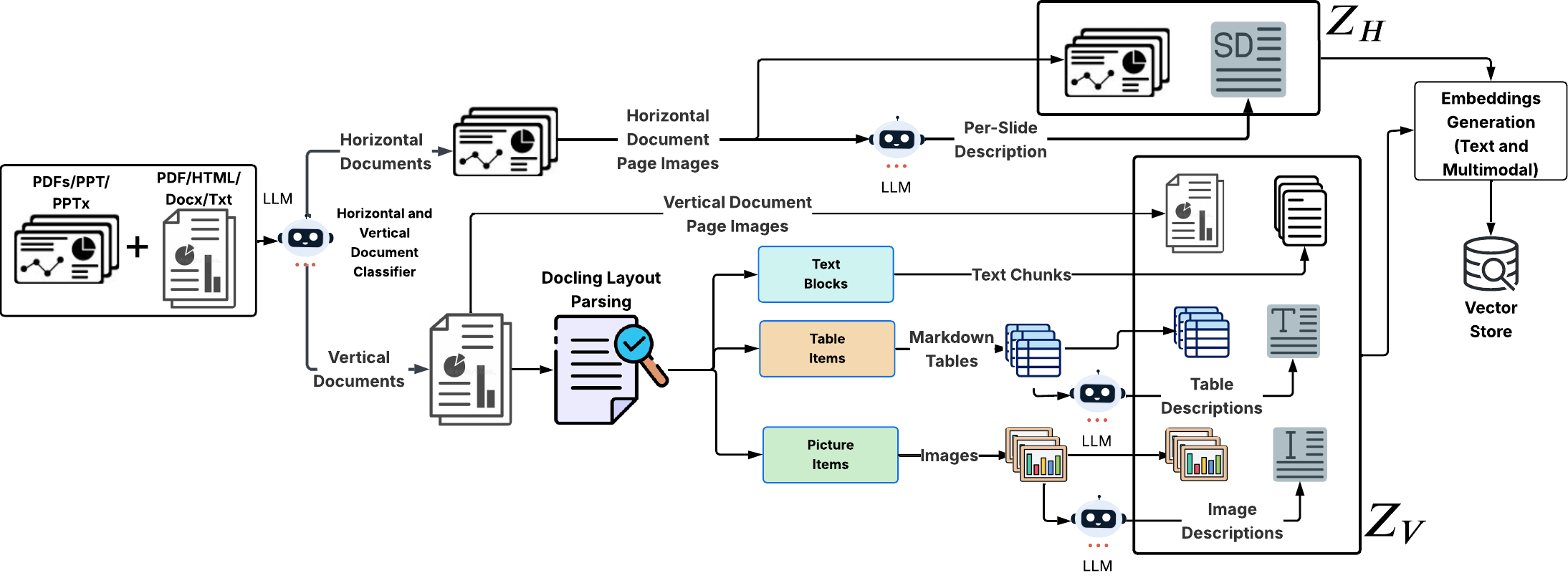}
\caption{
Overview of document structure-aware ingestion for vertically and horizontally structured enterprise documents. The pipeline adapts parsing and chunking strategies ($Z_H$ and $Z_V$) based on document structure. The realization of $Z_H$ and $Z_V$ for different variants is detailed in Figure \ref{fig:variants_chunking_ingestion}. 
}
\label{fig:structure-aware-ingestion}
\end{figure*}

\section{MM-BizRAG Methodology}
\label{sec:methodology}
In this section, we introduce MM-BizRAG. We first describe our document-structure-aware ingestion strategy (Section~\ref{subsec:SAI}), then present three MM-BizRAG variants (Section~\ref{sec:mmrag_variants}). In each variant, we specify one set of representations for retrieval and another for generator-ready context construction; these sets vary across variants. This adaptive production of distinct representations for the retriever and generator is central to MM-BizRAG.

\subsection{Document Structure-Aware Ingestion}
\label{subsec:SAI}
Let $\mathcal{D} = \{d_1, d_2, \dots, d_N\}$ denote a corpus of documents spanning multiple file formats. Each document may contain textual content, tables, figures, and a layout structure. We begin by assigning each document $d \in \mathcal{D}$ a structure label 
$s(d) \in \{V, H\}$ using an LLM-based classifier (or deterministically from file meta-data if such information exists), where $V$ denotes vertically 
structured documents ($D_V$) and $H$ denotes horizontally structured documents ($D_H$). 
This structure-aware partitioning determines the downstream representation pool 
construction strategy (Figure~\ref{fig:structure-aware-ingestion}).

\subsubsection{Vertical Document Representations} For each $d_v \in \mathcal{D}_V$, layout-aware parsing extracts aligned text blocks, tables, and intra-page pictures to construct the representations below.

\paragraph{Text Representation.} 
Text blocks from each parsed page $T_{d_v,i}$ where $i \in \{1, \ldots, |d_v|\}$ are concatenated into linearized representation $\mathcal{T}_{d_v} = \bigoplus_{i=1}^{|d_v|} T_{d_v,i}$ preserving reading order, with unique placeholders inserted at positions corresponding to tables and pictures as they appeared in the original document, to retain contextual alignment.

\paragraph{Table Representation.} Each table $k$ is converted to markdown $m_k$, then passed to an LLM to generate row-by-row description $s_k$. 
The collection of table representations is $\mathcal{R}^{tbl}_{d_v} = \{(m_k, s_k)\}_{k=1}^{|K_{d_v}|}$, where each $(m_k, s_k)$ is aligned to its corresponding placeholder in $\mathcal{T}_{d_v}$ via a positional pointer.

\paragraph{Picture Representation.} Each picture $p$ is processed by a VLM to generate description $s_p$ and filter uninformative content (logos, decorative elements). The collection of picture representations is $\mathcal{R}^{pic}_{d_v} = \{(p_j, s_{p_j})\}_{j=1}^{|P_{d_v}|}$, where each $(p_j, s_{p_j})$ is aligned to its corresponding placeholder in $\mathcal{T}_{d_v}$ via a positional pointer.

\paragraph{Page Images.} Full-page images are retained as the set $\Pi_{d_v} = \{\pi_{d_v,i}\}_{i=1}^{|d_v|}$.

\paragraph{Representation Pool.} The complete representation pool for $d_v$ is given by $\mathcal{R}_{d_v} = \{\mathcal{T}_{d_v}, \mathcal{R}^{tbl}_{d_v}, \mathcal{R}^{pic}_{d_v}, \Pi_{d_v}\}$, containing all extracted artifacts from $d_v$.

\subsubsection{Horizontal Documents Representations}
For $d_h \in \mathcal{D}_H$, explicit layout-aware parsing is not applied. Pages in horizontal documents, commonly presentation slides, are holistic semantic units where text, tables, pictures, and charts jointly convey page-level meaning, making fine-grained layout decomposition ineffective for complex presentation layouts. Unlike vertically structured documents, slides, often embed pictures within table cells, arrange content across free-form spatial zones sometimes unclear reading order, and carry meaning only at the level of the full page.

\paragraph{Page-level Representation Pool.} 
For each page $i$ in $d_h$, we extract page image $\pi_{d_h,i}$ and use a VLM to generate detailed textual description $\delta_{d_h,i}$ capturing all semantic content: salient text, visual elements, and their relationships. Crucially, $\delta_{d_h,i}$ is a comprehensive description encompassing every aspect of the page, so it could be considered as \textit{LLM based layout parsing}. The page-level representation is $(\delta_{d_h,i}, \pi_{d_h,i})$. The representation pool for $d_h$ is defined as $\mathcal{R}_{d_h} = \{(\delta_{d_h,i}, \pi_{d_h,i})\}^{|d_h|}_{i=1}$.

\subsubsection{Transformation Operators} 
Once $\mathcal{R}_{d_v}$ and $\mathcal{R}_{d_h}$ are constructed, we apply a document structure-specific transformation operator $Z_{s(d)}$ to transform the representation pools into a set of retrievable chunks $\mathcal{C}_d$: $\mathcal{C}_d = Z_{s(d)}(\mathcal{R}_{s(d)})$, where $s(d) \in \{V,H\}$.

The operator $Z_{s(d)}$ (i) composes a subset of representations from $\mathcal{R}_{s(d)}$ and (ii) segments them into chunks according to a chosen granularity. The chunk set $\mathcal{C}_d$ is embedded using either a text ($\mathcal{E}_t$) or multimodal ($\mathcal{E}_{mm}$) embedding model, or both.

\label{sec:mmrag_variants}
\begin{figure}[!h]
\centering
\includegraphics[width=\columnwidth]{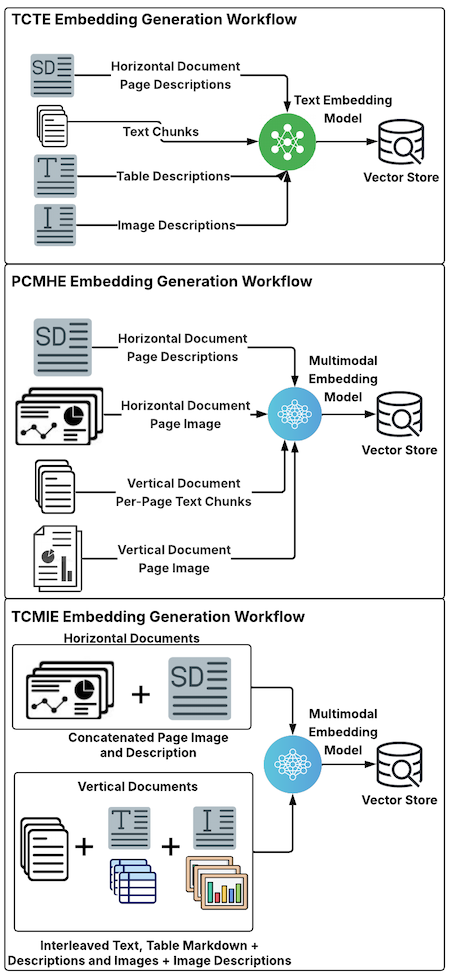}
\caption{
Overview of embedding generation strategies for different MM-RAG Variants.
}
\label{fig:variants_chunking_ingestion}
\end{figure}
\subsection{MM-BizRAG Variants} 
Based on the document structure-aware ingestion pipeline (Section \ref{subsec:SAI}), we explore three MM-BizRAG variants that share a common inference stack: query re-writer, list-wise LLM re-ranker~\cite{sun2023chatgpt}, and MM answer generator $G$. The inference pipeline is detailed in Appendix \ref{sec:inference_pipeline}. The variants differ along three dimensions: (i) how the transformation operator $Z_V$ or $Z_H$ constructs retrievable chunks from the representation pool $\mathcal{R}_{d_v}$ or $\mathcal{R}_{d_h}$, respectively, (ii) embedding models used for indexing and retrieval, and (iii) how retrieved chunks are assembled into generator-ready context, described by operator $\Phi$. Figure \ref{fig:variants_chunking_ingestion} details embedding generation for each variant. Each variant includes both textual and visual artifact representations (image for slide, pictures; markdown for tables) in generator context, as jointly conditioning on text and visual inputs improves grounding and reduces textual ambiguity \cite{wang2024comprehensivereviewmultimodallarge}. 

\subsubsection{Variant 1: Token level chunking Text embedding (TCTE)}\label{subsubsec:variant_1}
\paragraph{$\mathbf{Z}_{V}^{(1)}$: }

For $d_v \in \mathcal{D}_V$, $Z^{(1)}_V$ first composes $\mathcal{T}_{d_v}$, table descriptions $\{s_k\}_{k=1}^{|K_{d_v}|}$ from $\mathcal{R}^{tbl}_{d_v}$, and picture descriptions $\{s_p\}_{p=1}^{|P_{d_v}|}$ from $\mathcal{R}^{pic}_{d_v}$, all drawn from $\mathcal{R}_{d_v}$.
$\mathcal{T}_{d_v}$ is segmented using token based chunking respecting sentence boundaries, while each $s_k$ and $s_p$ forms a standalone chunk without segmentation. The final chunk set consists of segmented text chunks, table and picture description chunks.

\paragraph{$\mathbf{Z}_{H}^{(1)}$: }
For $d_h \in \mathcal{D}_H$, $Z^{(1)}_H$ selects slide-level descriptions $\{\delta_{d_h,i}\}_{i=1}^{|d_h|}$ from $\mathcal{R}_{d_h}$, where each $\delta_{d_h,i}$ forms a standalone chunk. The final chunk set consists of slide/page description chunks.

\paragraph{$\bm{\mathcal{E}}^{(1)}$: }
All chunks are embedded using text embedding model $\mathcal{E}_{t}$. Retrieval uses hybrid sparse-dense retrieval with reciprocal rank fusion (RRF). Retrieved chunks may originate from any of the chunk types described above.

\paragraph{$\mathbf{\Phi}^{(1)}$: }
Prior to generation, $\Phi^{(1)}$ constructs MM context as follows: (i) The retriever retrieves vertical document text chunks, table and/or picture description chunks. For each retrieved table or picture description chunk, we additionally identify the associated text chunk containing its placeholder. (ii) Within each text chunk, the table markdown and description $(m_k, s_k)$ and the picture artifact and description $(p, s_p)$ are injected at their original placeholder positions. (iii) For retrieved page-level chunks from $\mathcal{D}_H$, page image $\pi_{d_h,i}$ is concatenated to page description $\delta_{d_h,i}$ (see Figures \ref{fig:mm_reader_zoomed_in_view} and \ref{fig:chunk_type_handling_strategy}).

\subsubsection{Variant 2: Page level chunking Multimodal Page level hybrid embedding (PCMHE)}\label{subsubsec:variant_2}


\paragraph{$\mathbf{Z}_{V}^{(2)}$: }
For $d_v \in \mathcal{D}_V$, $Z^{(2)}_V$ constructs two chunk types per page: (i) page-text chunks by using the layout parsed page text $T_{d_v,i}$ (with table and image content replaced with placeholders), and (ii) page-image chunks from page images $\{\pi_{d_v,i}\}_{i=1}^{|d_v|}$.

\paragraph{$\mathbf{Z}_{H}^{(2)}$: }
For $d_h \in \mathcal{D}_H$, $Z^{(2)}_H$ follows $Z^{(1)}_H$ but constructs two chunk types per page from $\mathcal{R}_{d_h}$: (i) slide-text chunks using LLM generated slide-level descriptions $\{\delta_{d_h,i}\}_{i=1}^{|d_h|}$, and (ii) slide-image chunks from $\{\pi_{d_h,i}\}_{i=1}^{|d_h|}$, where each $\delta_{d_h,i}$ and $\pi_{d_h,i}$ forms a standalone chunk at slide granularity.

\paragraph{$\mathbf{\mathcal{E}}^{(2)}$: }
All chunks are embedded using $\mathcal{E}_{mm}$. Page-text and page-image retrieval are performed independently within the MM embedding space, and the resulting rankings are fused using RRF.

\paragraph{$\mathbf{\Phi}^{(2)}$: }
$\Phi^{(2)}$ constructs its context by pairing each retrieved vertical page-text chunk with its corresponding page image, and each horizontal page/slide-image chunk with its corresponding description. For pages from $\mathcal{D}_V$, table markdown $m_k$ is inserted alongside description $s_k$, and picture artifact $p$ alongside description $s_p$.

\subsubsection{Variant 3: Token level chunking Multimodal Interleaved embedding (TCMIE)}\label{subsubsec:variant_3}

\paragraph{$\mathbf{Z}_{V}^{(3)}$: }
For $d_v \in \mathcal{D}_V$, $Z_{V}^{(3)}$ selects $\mathcal{T}_{d_v}$, $\mathcal{R}^{tbl}_{d_v}$, and $\mathcal{R}^{pic}_{d_v}$ from $\mathcal{R}_{d_v}$. Like $Z_{V}^{(1)}$, it segments $\mathcal{T}_{d_v}$ into text chunks, but instead of treating tables and images as standalone chunks, $Z_{V}^{(3)}$ composes unified multimodal units by injecting each table markdown and description $(m_k, s_k)$ and picture artifact and description $(p_j, s_{p_j})$ at their corresponding placeholder positions within the respective text chunk.

\paragraph{$\mathbf{Z}_{H}^{(3)}$: } 
For $d_h \in \mathcal{D}_H$, $Z^{(3)}_H$ follows $Z^{(2)}_H$ but composes single unified multimodal chunks by combining each page description $\delta_{d_h,i}$ with its corresponding page image $\pi_{d_h,i}$, rather than treating them as separate chunks. Each page forms a single multimodal chunk $(\delta_{d_h,i}, \pi_{d_h,i})$ at page/slide granularity.

\paragraph{$\mathbf{\mathcal{E}}^{(3)}$: }
All chunks are embedded using $\mathcal{E}_{mm}$.

\paragraph{$\mathbf{\Phi}^{(3)}$: }
Since MM composition occurs during ingestion, $\Phi^{(3)}$ only performs transformations on retrieved chunks from $\mathcal{D}_V$. Each table/picture's content (markdown/image) is inserted alongside its textual description within the MM chunk unit.

\begin{figure*}[!th]
\centering
\includegraphics[width=\textwidth]{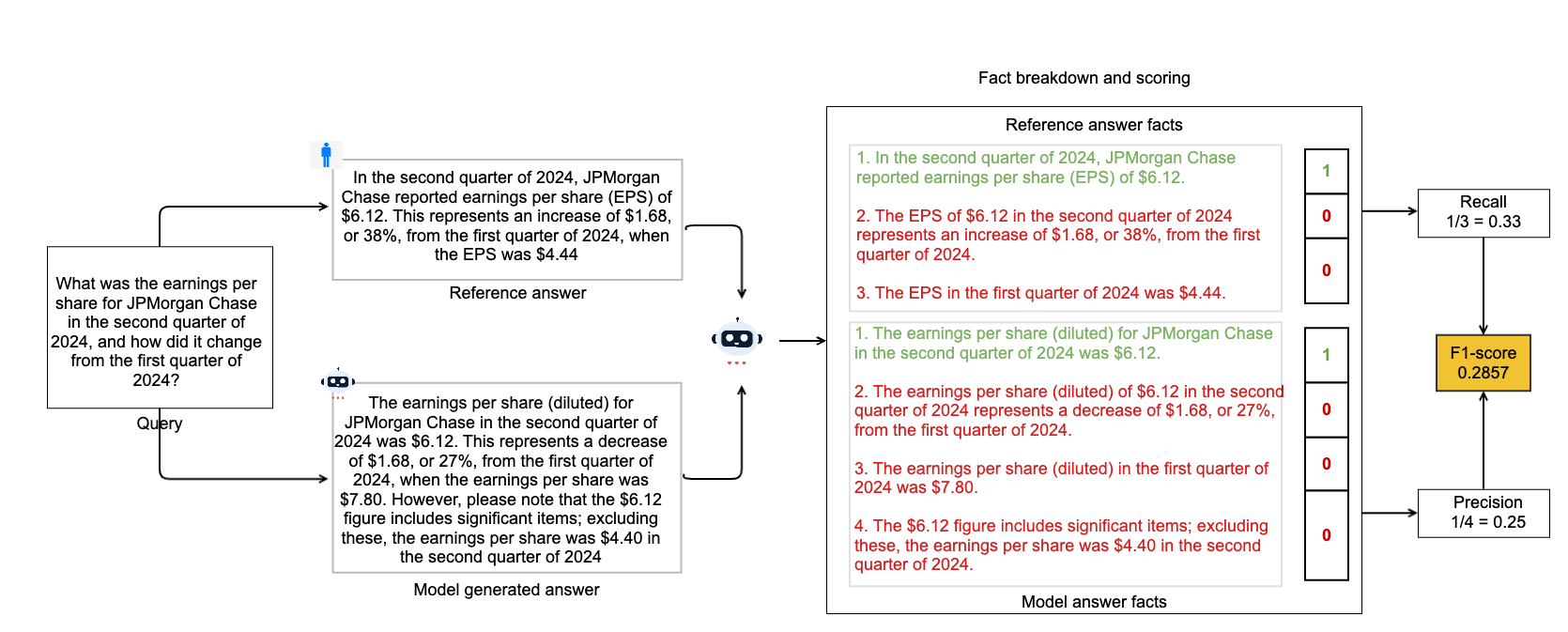}
\caption{
This figure presents an overview of the FastRAGEval metric. Given a query, a reference (ground truth) answer, and a model-generated answer, the metric decomposes each answer into a list of atomic facts. Each reference fact is checked for presence in the model-generated fact list to compute Recall, and each model-generated fact is checked for presence in the reference fact list to compute Precision. A fact scores 1 if present and 0 otherwise. The final F1-score is the harmonic mean of Precision and Recall.
}
\label{fig:llm_qa_metric}
\end{figure*}

\section{Experiment Setup}

\paragraph{Dataset.}
We evaluate our methods on an internal enterprise multimodal dataset and two public benchmarks. Our internal dataset includes varied enterprise documents—financial, legal, technical, and more—across HTML, PPTX, Word, and PDF formats. These files differ significantly in layout and modality, spanning 1908 questions across 1048 documents and 20429 pages. To assess generalizability, we additionally evaluate on SlideVQA~\cite{tanaka2023slidevqadatasetdocumentvisual} and FinRAGBench-V~\cite{zhao2025finragbenchvbenchmarkmultimodalrag}; details in Appendix \ref{sec:dataset}.

\paragraph{Baselines.}
Firstly, we compare with a text-only RAG baseline (e.g. \citet{zhao2024optimizing}) where we perform OCR on the text documents, apply basic sentence boundary preserving token wise chunking and embed the chunks using the same text embedding model $\mathcal{E}_t$. Then, we compare against two popular vision-centric MM-RAG pipelines, ColPali~\cite{faysse2024colpaliefficientdocumentretrieval} and VisRAG~\cite{yu2024visrag}. 

\paragraph{Models used.}
We use OpenAI text-embedding-3-large \cite{openai2024embeddings} as our text embedding model, cohere-embed-v4 \cite{cohere2025embeddings} and nomic-multimodal-embed-3b \cite{nomicembedmultimodal2025} for multimodal embeddings, Docling \cite{livathinos2025doclingefficientopensourcetoolkit} for vertical document layout parsing and GPT 4.1 family of models for other text generation tasks.\\
For other implementation details please refer to Appendix \ref{sec:impl_details}.

\subsection{Evaluation Metrics}
We prioritize recall-oriented metrics, well-suited for long-form enterprise Q\&A, and report metrics commonly used in prior work for each dataset.
For \textsc{SlideVQA}~\cite{tanaka2023slidevqadatasetdocumentvisual}, we report token recall (fraction of reference tokens in the generated answer). For \textsc{FinRAGBench-V}~\cite{zhao2025finragbenchvbenchmarkmultimodalrag}, we use the LLM-as-a-Judge binary metric from the original paper.

\label{sec:frageval}
\subsubsection{LLM as Judge Metrics}
\paragraph{FastRAGEval}
For our internal dataset with long-form answers, token match based QA metrics (EM, F1, BLEU, ROUGE) are ill-suited. RAGChecker~\cite{ru2024ragcheckerfinegrainedframeworkdiagnosing} decomposes reference answers into atomic claims to compute claim-level metrics (precision, recall, F1), but its two-call LLM pipeline increases cost, latency, and error propagation. We introduce \textbf{FastRAGEval (FRE)}, a single-call, reference-based LLM-as-a-Judge suite (Fig.~\ref{fig:llm_qa_metric}) that extracts key facts and returns precision, recall, and F1 in one LLM call. For our experiments we mainly use FRE recall. To validate FRE against human judgment, we collect annotations on 200 instances (100 per system: ColPali and VisRAG) from two independent annotators using a 3-point rubric (0: Incorrect, 1: Partially Correct, 2: Completely Correct), stratified across modalities and document types, conducted blind to the generating system.

\paragraph{Faithfulness}
We also evaluate using an internally developed faithfulness metric (similar to \citet{ragas2024}), comparing generated answers against (multimodal) retrieved sources.

\begin{table*}[!th]
\resizebox{\textwidth}{!}{%
\begin{tabular}{l|ccccccccc}
\hline
\multirow{2}{*}{\textbf{RAG   Pipeline}} &
  \multicolumn{3}{c|}{\textbf{SlideVQA}} &
  \multicolumn{3}{c|}{\textbf{FinRAGBench-V}} &
  \multicolumn{3}{c}{\textbf{Internal   Dataset}} \\ 
 &
  \multicolumn{1}{c}{RC} &
  FRE & Latency &
  \multicolumn{1}{|c}{RC} &
  FRE & Latency &
  \multicolumn{1}{|c}{RC} &
  FRE & Latency\\ \hline
  \multicolumn{10}{l}{\textit{Baselines}} \\
\quad Text-Only            & 66.1 & 67.8 & 5.2 &
\multicolumn{1}{|c}{54.84} & 60.3 & 6.4      & \multicolumn{1}{|c}{76.95}  & 83.7 & 7.6   \\ 
\quad ColPali              & 75.2 & 83.6  & -     & \multicolumn{1}{|c}{45.7} & 49.3  & -    & \multicolumn{1}{|c}{-}   & - &       -       \\ 
\quad VisRAG              & \multicolumn{1}{c}{71.3} & 78.8 & -  & \multicolumn{1}{|c}{44.1} & 46.0   & - & \multicolumn{1}{|c}{-}           & -       & -       \\ 
\hline
\multicolumn{10}{l}{\textit{Our Approach}} \\
\quad TCTE (OAI v3-large)          & \multicolumn{1}{c}{86.0} & 87.3   & 34.3    & \multicolumn{1}{|c}{75.0} & {80.2} & 11.9 & \multicolumn{1}{|c}{{ 82.0}} & \textbf{88.1} & 11.1\\ 
\quad PCMHE   (Nomic $\mathcal{E}_{mm}$) &
  \multicolumn{1}{c}{\textbf{89.1}} &
  \textbf{89.9} & 28.4 &
  \multicolumn{1}{|c}{{75.0}} &
  79.6 & 21.2 &
  \multicolumn{1}{|c}{81.6} &
  87.6 & 18.0\\
\quad PCMHE   (Cohere $\mathcal{E}_{mm}$) &
  \multicolumn{1}{c}{{87.7}} &
  89.06 & 28.1 &
  \multicolumn{1}{|c}{\textbf{77.8}} &
  \textbf{82.4} & 22.7 &
  \multicolumn{1}{|c}{\textbf{82.2}} &
  87.8 & 22.0\\ 
\quad TCMIE   (Cohere $\mathcal{E}_{mm}$) & \multicolumn{1}{c}{87.4} & 88.2 & 29.5 & \multicolumn{1}{|c}{70.8} & 76.9      & 11.3  & \multicolumn{1}{|c}{81.0} & 88.0  & 11.3   \\ 
\quad TCMIE   (Nomic $\mathcal{E}_{mm}$)  & \multicolumn{1}{c}{86.0} & 87.2   & 28.8     & \multicolumn{1}{|c}{63.2} & 71.0  &  7.3   & \multicolumn{1}{|c}{78.0}       & 85.0 & 10.2          \\ \hline
\end{tabular}%
}
\caption{Overall results across SlideVQA, FinRAGBench-V, and our Internal Dataset. RC (RAGChecker) and FRE (FastRAGEval) metrics report recall. Latency reports the mean end-to-end inference pipeline latency in seconds. \textbf{Bold} values indicate the best-performing system for each dataset–metric pair. Nomic $\mathcal{E}_{mm}$:  nomic-multimodal-embed-3b; Cohere $\mathcal{E}_{mm}$: cohere-embed-v4;  OAI v3-large: OpenAI text-embedding-3-large.}
\label{tab:overall_results}
\end{table*}

\section{Results and Discussion}
\subsection{MM-BizRAG System Evaluations}
Table \ref{tab:overall_results} presents an overall comparison between the MM-BizRAG variants, ColPali, VisRAG, and a simple text-only RAG baseline across \textsc{SlideVQA} and \textsc{FinRAGBench-V}.

On \textsc{SlideVQA}, MM-BizRAG variants outperform ColPali and VisRAG on both RagChecker-Recall and FRE-Recall metrics. The range of improvement on the FRE-Recall metric with respect to ColPali is \textbf{3.6-6.3\%}. The range of improvement with respect to VisRAG is even more: \textbf{8.4-11.1\%}. The difference in RagChecker-Recall scores are also in the similar range. SlideVQA's visually-integrated layouts explain why ColPali and VisRAG outperform the text-only baseline. MM-BizRAG nonetheless achieves the strongest results by integrating both textual and visual representations, showing the importance of text representations even for slides where vision-centric pipelines should shine.

\begin{figure}[t]
\centering
\includegraphics[width=0.98\columnwidth]{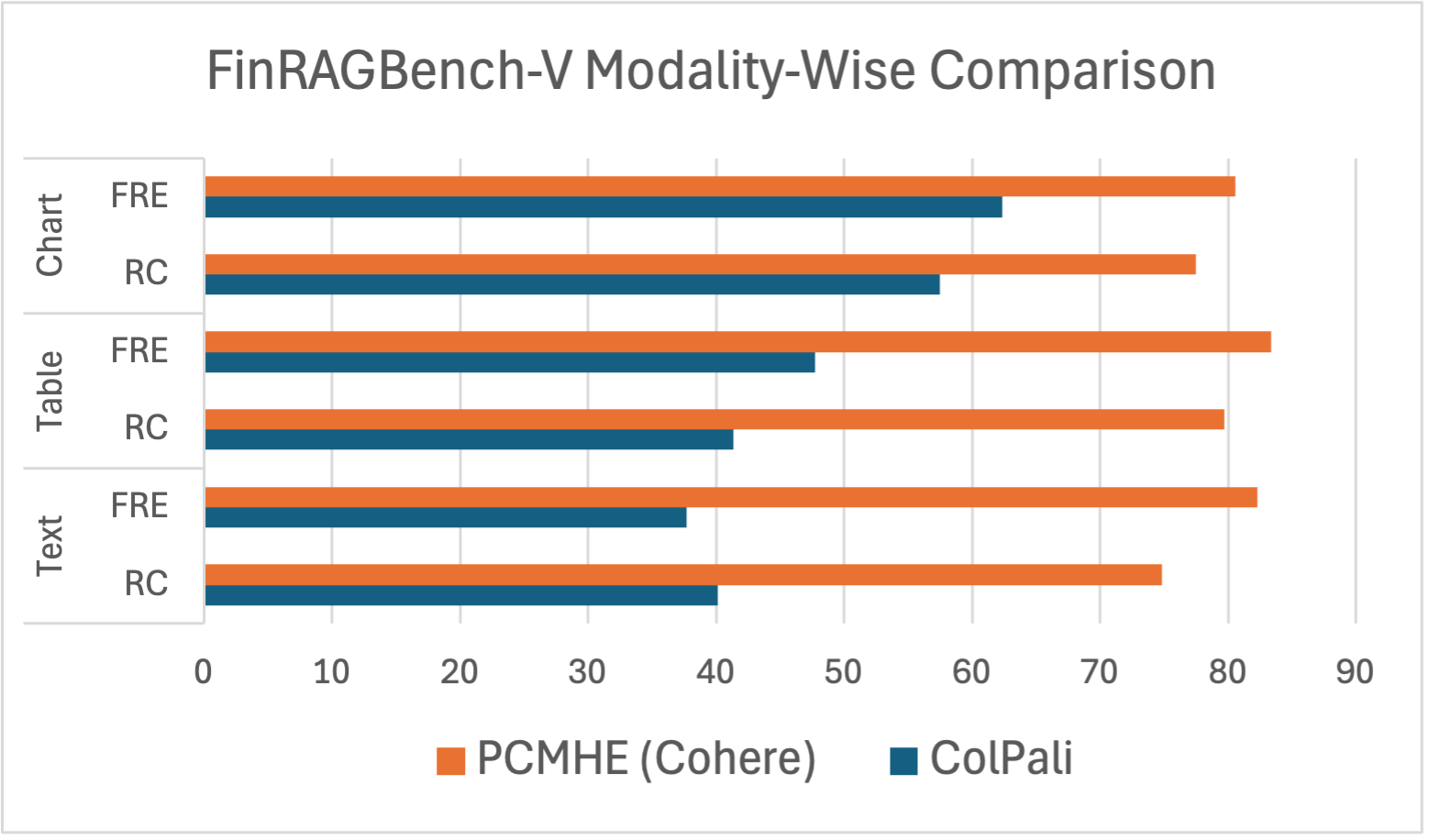}
\caption{
Modality-wise comparison of recall metrics between PCMHE (Cohere $\mathcal{E}_{mm}$) and ColPali on FinRAGBench-V. PCMHE outperforms ColPali across all the modalities by a substantial margin.
}
\label{fig:modality_bar_chart}
\end{figure}

On \textsc{FinRAGBench-V}, MM-BizRAG variants substantially outperform ColPali and VisRAG by more than \textbf{25\%} (up to \textbf{32\%} for best variant) on both RC-Recall and FRE-Recall. This dataset consists of vertical documents containing a high density of text, structured tables, and charts with cross-page discourse dependencies. Methods that depend solely on page image-centric MM-RAG degrade significantly for such scenarios (Figure \ref{fig:modality_bar_chart}), demonstrating the importance of layout aware parsing with LLM driven artifact transformation. In fact, the text-only RAG baseline can outperform vision centric pipelines for such datasets. MM-BizRAG further improves on this baseline with much higher performance across all modalities.

On the internal dataset with even more complex business documents, we compare only against our text-only RAG baseline due to model weight access and data privacy constraints. Across both the recall metrics, all MM-BizRAG variants outperform the text-only RAG as shown in Table \ref{tab:overall_results}. While the text-only baseline remains competitive on text-based questions, it exhibits clear degradation on table and picture based questions (Table \ref{tab:internal_detailed_perf_numbers}). In contrast, MM-BizRAG maintains a more balanced performance across different modalities.

Faithfulness scores \textbf{(>90\%)} (Appendix \ref{sec:detailed_results}) for all MM-BizRAG variants across all datasets further indicates that our designs generate answers that are consistently supported by retrieved evidence, reducing unsupported or hallucinated claims.

The majority of pipeline latency (75–80\%) is attributable to GPT-4.1 generation under experimental environment rate limits, and would decrease under production-tier TPM allocations. For vertical documents, PCMHE's page-image representations nearly double latency relative to TCTE with only marginal recall gains (for horizontal they are more comparable). TCTE strikes the most favorable performance–latency trade-off, achieving recall within 1–3\% of the best configuration while incurring roughly half the latency of PCMHE on average, \textbf{making it our recommended configuration for production deployment}.

\subsection{FastRAGEval vs. RAGChecker}
FRE-Recall aligns more strongly with human judgements than RC across all three correlation measures, as validated by the annotation process described in Section~\ref{sec:frageval}. Pearson's $r$: \textbf{0.808 vs. 0.748}, Spearman's $\rho$: \textbf{0.808 vs. 0.736}, and Kendall's $\tau_b$: \textbf{0.808 vs. 0.725}. The two independent annotators achieve a high inter-annotator agreement measured via Cohen's kappa of \textbf{0.966}.

\section{Conclusion}
We presented MM-BizRAG, a multimodal RAG framework that re-centers the importance of explicit document structure and artifact-aware parsing for enterprise question answering. By proactively extracting and representing the rich, structured information in complex documents -- through layout-aware parsing for reports and holistic page-level representations for slides -- MM-BizRAG enables higher quality e2e performance than approaches relying solely on page-level image representations.
Our experiments on three large, heterogeneous datasets demonstrated that MM-BizRAG consistently outperforms SOTA vision-centric baselines, with especially strong gains on vertically structured documents. 
We further introduced FastRAGEval, a single-call LLM-Judge metric for fine-grained generative recall, which aligns more closely with human judgment than previous approaches while still being more efficient.
Overall, our findings re-establish the value of direct, structure-aware document understanding in multimodal RAG, and provide a practical blueprint for harmonizing parsing-based and embedding-based approaches in real-world enterprise QA. 



\section{Limitations}
While MM-BizRAG demonstrates strong performance across enterprise documents, our study is subject to limitations. Our public benchmark evaluation on slide-type documents is primarily based on the \textsc{SlideVQA} benchmark, which, although useful, consists of relatively simple slides and does not fully reflect the complexity and diversity of industry-grade presentations; we were unable to include more challenging datasets such as \textsc{Real-MM-RAG} or our internal dataset, which would provide a more rigorous test of our pipeline. Additionally, our experiments are limited to two public datasets, selected to represent both vertically and horizontally structured documents. This focus allowed us to analyze the impact of structure-aware ingestion, but a broader evaluation on more diverse and complex enterprise documents would be valuable for assessing even more robust generalization. For \textsc{FinRAGBench-V}, we were only able to process a subset of 213 English-language PDFs the authors released in PDF file format, rather than the full corpus of over 1,100 documents which were released as binary part files, due to compute constraints. As a result, our findings may not fully capture the pipeline’s performance across the entire benchmark, and we did not evaluate multilingual capabilities, which are often important in enterprise settings. Furthermore, we compared MM-BizRAG against only two open-source baselines, Colpali and VisRAG, as these were the most suitable and available for our evaluation scenario; including a wider range of baselines, especially more recent or proprietary systems, would provide a more comprehensive comparison. Finally, due to privacy and organizational constraints, we are unable to release our proprietary enterprise dataset at this time, though we are actively exploring options for releasing an anonymized or synthetic version in the future to support reproducibility and further research.

\section*{Disclaimer}
This paper was prepared for informational purposes in part by the Machine Learning Center of Excellence group of JPMorgan Chase \& Co. and its affiliates ("JP Morgan”) and is not a product of the Research Department of JP Morgan. JP Morgan makes no representation and warranty whatsoever and disclaims all liability, for the completeness, accuracy or reliability of the information contained herein. This document is not intended as investment research or investment advice, or a recommendation, offer or solicitation for the purchase or sale of any security, financial instrument, financial product or service, or to be used in any way for evaluating the merits of participating in any transaction, and shall not constitute a solicitation under any jurisdiction or to any person, if such solicitation under such jurisdiction or to such person would be unlawful.

\bibliography{custom}

\appendix

\section{Appendix: Dataset Details}
\label{sec:dataset}

\subsection{Internal Enterprise Dataset.} Our internal dataset consists of real-world enterprise documents spanning multiple lines of business, including financial filings and earnings reports, presentation slide decks, meeting notes, legal documents, press releases, technical documentation, business policies, employee manuals and investment insights. Documents span multiple file formats, including HTML, PPTX, Word Documents, and PDFs, and exhibit substantial variation in layout structure and modality usage. This diversity reflects the heterogeneity typically encountered in enterprise knowledge systems which is not commonly found in popular MM-RAG datasets such as ViDoRe-v2~\cite{mace2025vidore}, LongDocURL~\cite{deng2025longdocurl}, and MMLongBenchDoc~\cite{ma2024mmlongbench}. 

The questions were authored by subject matter experts and annotators. The questions span factual, analytical, and procedural topics. Additionally, the dataset includes modality labels for each question, indicating whether the answer relies on text, tables, images, or combination thereof, and requires aggregating information across multiple documents and multiple pages. The dataset contains 1908 questions across 1048 documents consisting a total of 20429 pages. Statistics of internal dataset are shown in Table \ref{tab:support_summary} and \ref{tab:internal_dataset_file_type_counts}.

\begin{table}[h!]
\centering
\resizebox{\columnwidth}{!}{%
\begin{tabular}{llr}
\hline
\multicolumn{3}{c}{\textbf{Internal Dataset}} \\
\hline
\textbf{Category} & \textbf{Values} & \textbf{Support} \\
\hline
\multirow{5}{*}{\textbf{Modality}}
    & all      & 1908 \\
    & text     & 941 \\
    & table    & 444 \\
    & picture  & 451 \\
    & mixture  & 60  \\
\hline
\multirow{5}{*}{\textbf{File type}}
    & pdf      & 1570 \\
    & ppt      & 152  \\
    & docx     & 35   \\
    & html     & 86   \\
    & txt      & 51   \\
\hline
\multirow{2}{*}{\textbf{Document type}}
    & document & 1588 \\
    & slide    & 307  \\
\hline
\multirow{3}{*}{\textbf{Answer type}}
    & Numeric              & 244  \\
    & Non-numeric          & 1198 \\
    & Numeric + Non-numeric& 466  \\
\hline
\end{tabular}%
}
\caption{QA Support by Modality, File Type, Document Type, and Answer Type}
\label{tab:support_summary}
\end{table}

\begin{table}[!h]
\centering
\resizebox{\columnwidth}{!}{%
\begin{tabular}{lr}
\hline
\multicolumn{2}{c}{\textbf{Internal Dataset}} \\
\hline
\textbf{File Type} & \textbf{Number of Documents} \\
\hline
PDFs  & 271 \\
PPTX  & 26  \\
DocX  & 18  \\
HTML  & 15  \\
TXT   & 718 \\
\hline
\textbf{Total Documents} & \textbf{1048} \\
\textbf{Total Pages} & \textbf{20429} \\
\hline
\end{tabular}%
}
\caption{Number of Documents by File Type}
\label{tab:internal_dataset_file_type_counts}
\end{table}

\subsection{Public Benchmarks.} To assess competitiveness of our methods with prior work while fostering reproducibility (as we are currently unable to release our internal dataset), we additionally evaluate on SlideVQA~\cite{tanaka2023slidevqadatasetdocumentvisual} and FinRAGBench-V~\cite{zhao2025finragbenchvbenchmarkmultimodalrag}. SlideVQA includes presentation-style documents which explicitly evaluates our horizontal document ingestion method. FinRAGBench-V is vital for our analysis due to two unique features: its use of documents with dense report-like layouts (vertical documents), and its support for evaluating over text, table and picture modalities. This makes it particularly well-suited for testing our core hypothesis whether vision only pipelines can completely replace careful layout parsing, especially on vertically structured documents. Statistics of SlideVQA and FinRAGBench-V datasets are shown in Table \ref{tab:public_support_questions} and \ref{tab:public_support_docs}.

\begin{table}[!h]
\centering
\resizebox{\columnwidth}{!}{%
\begin{tabular}{ccr}
\hline
\textbf{Dataset} & \textbf{Modality} & \textbf{Support} \\
\hline
\multirow{1}{*}{SlideVQA}
    & - & 1652 \\
\hline
\multirow{4}{*}{FinRAGBench-V} 
    & all      & 539 \\
    & text     & 144 \\
    & table    & 216 \\
    & picture  & 156 \\
\hline
\end{tabular}%
}
\caption{Number of questions by modality in FinRAGBench-V and SlideVQA}
\label{tab:public_support_questions}
\end{table}

\begin{table}[!h]
\centering
\resizebox{\columnwidth}{!}{%
\begin{tabular}{cccc}
\hline
\textbf{Dataset} & \textbf{Number of Documents} & \textbf{Number of Pages} \\
\hline
\multirow{1}{*}{SlideVQA}
    & 300 & 60000 \\
\hline
\multirow{1}{*}{FinRAGBench-V} 
    & 213      & 11432 \\
\hline
\end{tabular}%
}
\caption{Number of documents and pages in FinRAGBench-V and SlideVQA}
\label{tab:public_support_docs}
\end{table}

\section{Appendix: Implementation Details}
\label{sec:impl_details}
Summary of the models used across different components are given in Table \ref{tab:component_models}.

\paragraph{Document Parsing.} 
We use Docling~\cite{livathinos2025doclingefficientopensourcetoolkit} for layout-aware parsing of vertically structured documents in MM-BizRAG. Within Docling~\cite{livathinos2025doclingefficientopensourcetoolkit}, text-bearing elements—paragraphs, section headers, page headers/footers, list items, titles, and formulae—are extracted from our proprietary corpus and FinRAGBench using EasyOCR \cite{jaidedai2020easyocr} and PyPdfium2 \cite{pypdfium2}. 

For the text-only baseline, we disable layout-aware parsing and extract text from scanned and born-digital PDFs with EasyOCR \cite{jaidedai2020easyocr} and PyPdfium2 \cite{pypdfium2} across all datasets.

\paragraph{Table-to-Markdown Conversion.}
Tables detected in the layout are converted into structured markdown using Tableformer \cite{nassar2022tableformer}.

\paragraph{LLM Generated Descriptions. } All generative descriptions used in ingestion, including table markdown to description, picture to description, and horizontal page to text description - are produced using the GPT-4.1 family of models.

\paragraph{Answer Generator model $G$.} For answer generation in both baselines (Text-Only, ColPali, and VisRAG) and all MM-BizRAG variants, we use GPT-4.1 as the shared generative backbone to ensure that answer quality is not biased due to different answer generation models.

\paragraph{Baseline Embedding Models.} 
As part of the VisRAG implementation, we use VisRAG-Ret \cite{yu2024visrag} as the embedding model and for the ColPali implementation, we use colpali-v1.3-hf embedding model.

\paragraph{Baseline Prompts.} 
The prompts for ColPali and VisRAG are adopted directly from \citet{yu2024visrag} to ensure canonical baseline performance.

\paragraph{Text Embedding Model.} All textual chunks across the variants are embedded using OpenAI's text-embedding-3-large model.

\paragraph{Multimodal Embedding Model.} For MM retrieval, we evaluate with two MM embedding models: nomic-multimodal-embed-3b from the Nomic Embed suite and cohere-embed-v4 model.

\paragraph{LLM-as-a-Judge Evaluators.} 
For FinRAGBench binary LLM judge, we follow the original setup in \citet{zhao2025finragbenchvbenchmarkmultimodalrag} which uses the GPT-4o model. For RAGChecker \cite{ru2024ragcheckerfinegrainedframeworkdiagnosing} and FastRAGEval (described in Section \ref{sec:frageval}), we use the same GPT-4.1 model.

\begin{table*}[!h]
\resizebox{\textwidth}{!}{%
\begin{tabular}{c|cc}
\hline
\textbf{Pipeline} & \textbf{Component}                   & \textbf{Model}           \\ 
\hline
\multirow{7}{*}{Ingestion} & Horizontal and Vertical Document Labelling & gpt-4.1-mini-2025-04-14 \cite{openai2025gpt41}  \\ 
& Layout-Aware Parsing & Docling \cite{livathinos2025doclingefficientopensourcetoolkit} \\ 
& OCR Text Extraction & EasyOCR \cite{jaidedai2020easyocr}\\ 
                  & Horizontal Page Image to Description & gpt-4.1-2025-04-14 \cite{openai2025gpt41}       \\ 
                  & Table Markdown to Description        & gpt-4.1-mini-2025-04-14  \\ 
                  & Picture to Description               & gpt-4.1-mini-2025-04-14  \\ 
                  & Text Embedding                       & text-embedding-3-large \cite{openai2024embeddings} \\ 
                  & Nomic Multimodal Embedding           &  nomic-multimodal-embed-3b \cite{nomicembedmultimodal2025}                        \\
                  & Cohere Multimodal Embedding          &  cohere-embed-v4                        \\ \hline
\multirow{3}{*}{Inference} & Query Rewriter                             & gpt-4.1-mini-2025-04-14 \\ 
                  & LLM List-Wise Re-ranker              & gpt-4.1-mini-2025-04-14  \\ 
                  & LLM Answer Generator                           & gpt-4.1-2025-04-14       \\ \hline
\multirow{3}{*}{LLM-as-a-Judge} & FinRAGBench Binary Judge \cite{zhao2025finragbenchvbenchmarkmultimodalrag}                             & gpt-4o-2024-05-13 \\ 
                  & RagChecker \cite{ru2024ragcheckerfinegrainedframeworkdiagnosing}              & gpt-4.1-2025-04-14  \\ 
                  & FastRAGEval                         & gpt-4.1-2025-04-14       \\ \hline
\multirow{3}{*}{Baseline Models} & ColPali \cite{faysse2024colpaliefficientdocumentretrieval}                             & colpali-v1.3-hf\cite{faysse2024colpaliefficientdocumentretrieval} \\ 
                  & VisRAG \cite{yu2024visrag}             & MiniCPM-V 2.0 (openbmb/VisRAG-Ret)  \\ 
 \hline
\end{tabular}%
}
\caption{Overview of models used for each component in our ingestion inference pipeline.}
\label{tab:component_models}
\end{table*}

\section{Appendix: Classification Accuracy of the Vertical-Horizontal Document Classifier}
To measure the accuracy of the vertical-horizontal document classifier a dataset of 517 documents was created, consisting of 299 horizontal documents and 218 vertical documents. The performance results of the classifier are listed in the Table \ref{tab:vertical-horizontal-doc-classifier-results}

\begin{table}
    \centering
    \begin{tabular}{l|l} \hline
         \textbf{Metric} & \textbf{Score} \\ \hline
        Precision & 100.00 \\
        Recall    & 83.28 \\
        F1-score  & 90.87\\ \hline
    \end{tabular}
    \caption{Classification performance of the vertical-horizontal document classifier in the ingestion pipeline.}
    \label{tab:vertical-horizontal-doc-classifier-results}
\end{table}

\section{Appendix: Inference Pipeline }
\label{sec:inference_pipeline}
\begin{figure*}[!h]
\centering
\includegraphics[width=2.2\columnwidth]{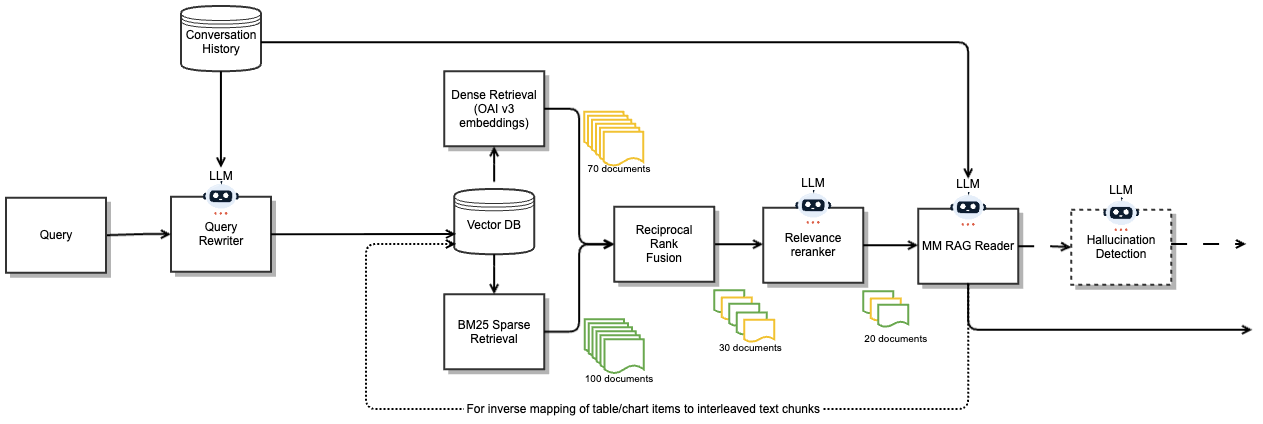}
\caption{Overview of the answer generation pipeline}
\label{fig:inference_pipeline_overall}
\end{figure*}

In this section we describe our overall inference pipeline approach for the text-centric retrieval with inference-time MM assembly (Variant 1, \ref{subsubsec:variant_1}) ingestion pipeline. 
Figure \ref{fig:inference_pipeline_overall}, describes our overall pipeline, the query rewriter (for the query rewriter prompt refer to Figure \ref{fig:query_rewriter_prompt}), rewrites the user query in the context of the conversation history. The rewritten query is passed to a hybrid retrieval pipeline which retrieves 70 document chunks using cosine similarity search, on dense embeddings and 100 document chunks based on BM25 search the two sets of retrieved document chunks are combined using Reciprocal Rank Fusion (RRF) and the top 30 chunks from the resulting list are passed onto the reranking stage. We perform an LLM-based listwise reranking (for the reranker prompt refer to Figure \ref{fig:document_reranking_prompt}) of the 30 document chunks and select the top 20 reranked document chunks which are sent to the multimodal VLM for answer generation.\\

We experimented with various types of LLM-based re-rankers \cite{sun2023chatgpt}, ultimately selecting the list-wise re-ranker (referred to as the permutation generation-based re-ranker in \cite{sun2023chatgpt}). We compared list-wise re-ranking to point-wise re-ranking, where the LLM is provided with an individual chunk and the query, and is asked to output a score from 0 to 10 based on a predefined guideline. On our benchmark datasets, we consistently found the end-to-end performance of the pipeline containing the point-wise re-ranking to be 2–3\% points better. However, this approach requires as many LLM calls as there are L1-retrieved chunks (30 in our case), resulting in significantly higher latency. While multi-threading the API calls can mitigate this to some extent, practical limitations arise due to API rate limiting, which throttles requests if we attempt to process all 30 queries simultaneously(in practice, our API throttles with more than 4–6 concurrent threads).

Furthermore, unlike the permutation generation approach described in \cite{sun2023chatgpt}, we do not need to implement any complex, sequentially dependent windowing logic. Thanks to our long context window (theoretically up to 1 million tokens, though we never use more than 10\% of this), we require at most a single API call for re-ranking.
Finally, we hypothesize that as models continue to improve in their ability to attend to long context sequences, list-wise ranking - which presents the re-ranker with all passages simultaneously - can theoretically capture more salient global relevance linkages across passages. This may affect the local relevance score of individual passages. For example, a passage that appears less relevant in isolation may become important in the presence of another highly relevant chunk (e.g., from the same file), thereby contributing to a more complete and accurate answer.

\begin{figure*}[!h]
\centering
\includegraphics[width=2.2\columnwidth]{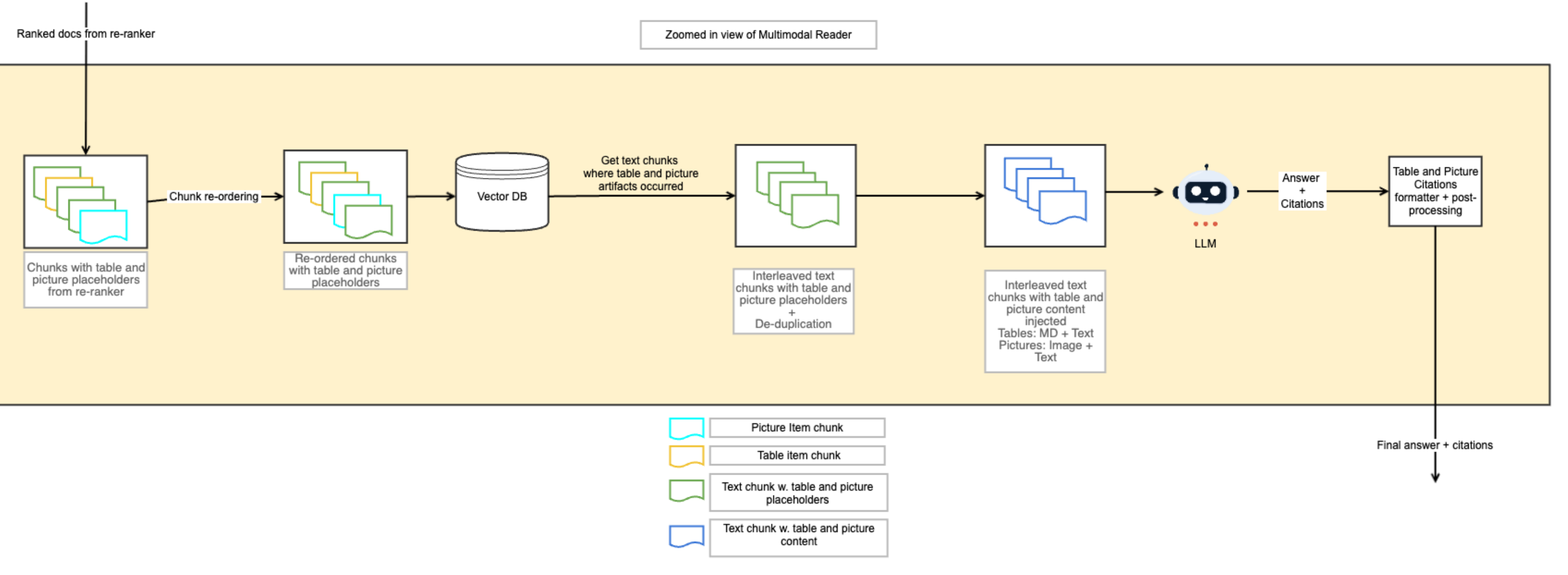}
\caption{Answer generation workflow for Variant 1 (Section 2.2.1). Retrieved text, table, and picture chunks are reordered by document order inspired by~\cite{yu2024defenserageralongcontext}; for table and picture chunks, the vector store is queried to retrieve their aligned parent text chunk, artifacts are injected at their placeholder positions, chunks are de-duplicated, and the resulting interleaved text-image-table representation is sent to the LLM for answer generation.}
\label{fig:mm_reader_zoomed_in_view}
\end{figure*}

\begin{figure*}[!h]
\centering
\includegraphics[width=2.2\columnwidth]{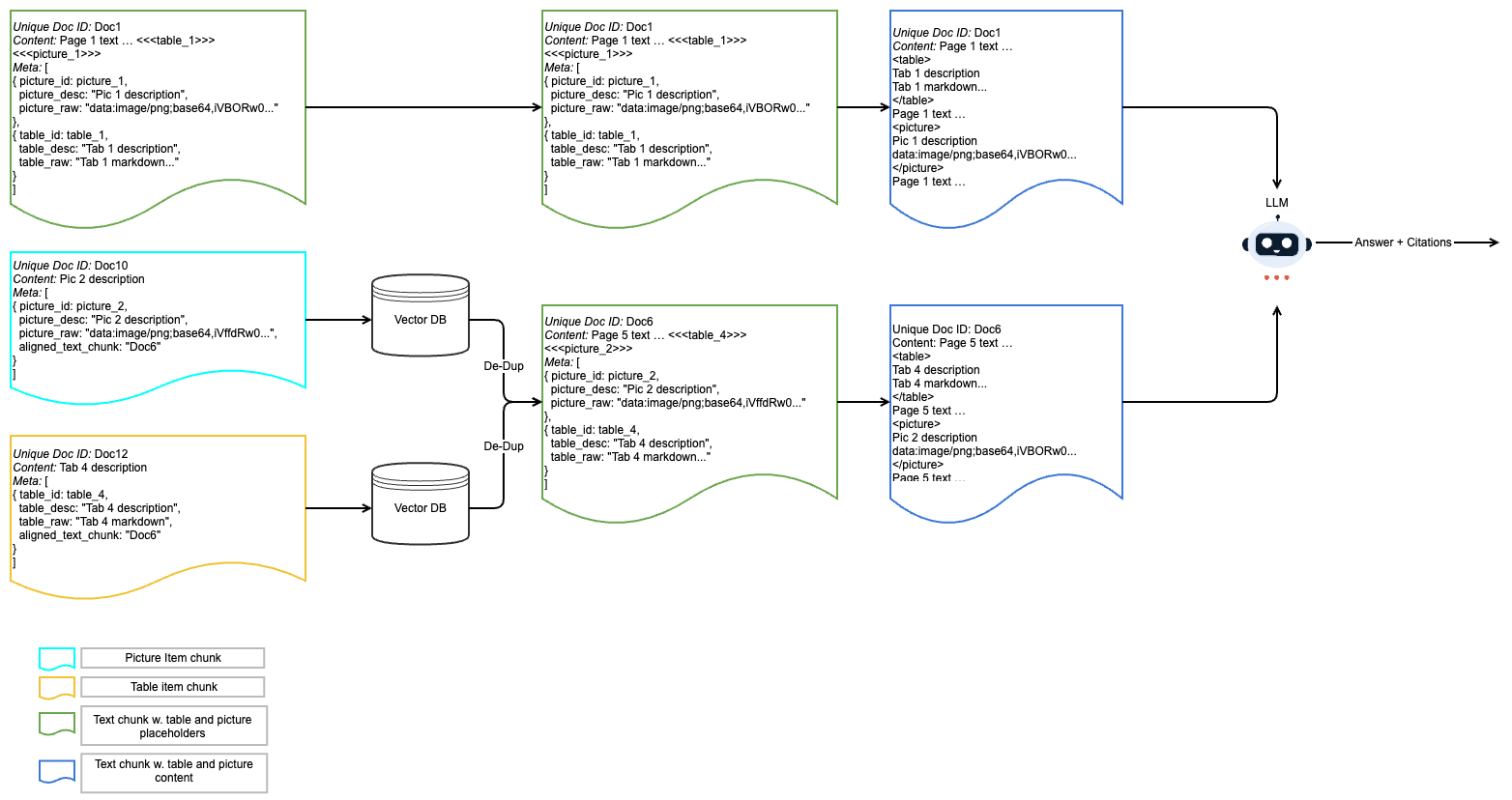}
\caption{Preprocessing steps for the three chunk types (text: green, picture: blue, table: yellow). For text chunks, \texttt{<<<picture\_i>>>} and \texttt{<<<table\_i>>>} placeholders are replaced with the base64 image and markdown respectively. For picture/table chunks, the aligned text chunk is retrieved from the vector store and artifacts are injected at the appropriate placeholder, producing interleaved text-image-table chunks (blue) sent to the LLM for answer generation. 
}
\label{fig:chunk_type_handling_strategy}
\end{figure*}

Figure \ref{fig:mm_reader_zoomed_in_view} and Figure \ref{fig:chunk_type_handling_strategy} provide a detailed view of our multimodal answer generator. The generator component obtains the re-ranked document chunks from the re-ranker which can be one of text chunk, image chunk (which contains a base 64 representation of an image and its LLM generated description) and, table chunks (which contains the markdown of a table and its LLM generated description) or in the case of horizontal documents the slide image and description chunks.  First, inspired from \cite{yu2024defenserageralongcontext}, the text chunks from the same documents are re-ordered to be in the same order as they appear in the original document. Next, for text chunks we inject the image and table artifacts belonging to the particular chunk from its meta-data and inject them to their corresponding text chunk and for image and table chunks we retrieve the corresponding text chunk that they are a part of (from the vector DB) and inject these artifacts into their text chunk, this results in an interleaved text, image and table representation for each chunk. The chunks are then de-duplicated and the prompt is constructed to be sent to the answer generation LLM.

During the prompt construction phase, each chunk is first mapped to a simple id, which is the index of chunks in the list of document chunks, this is done to avoid the LLM from having to replicate the complex ids for chunks in its citations and to reduce chances for hallucinations. Next we add the chunks to the prompts and we add the chunk id at the start and end of each chunk based on the technique from \cite{li2024retrievalaugmentedgenerationlongcontext}. Then, we add any custom user instructions to the prompt and finally we add instructions to the prompt for interleaving the answers and its citations. The constructed prompt is then sent to the VLM for answer generation.

\clearpage
\section{Appendix: Detailed Results}
\label{sec:detailed_results}
\begin{table*}[!h]
\centering
\resizebox{0.98\textwidth}{!}{%
\begin{tabular}{ll|ccc}
\hline
\multicolumn{5}{c}{\textbf{Internal Dataset}} \\
\hline
\textbf{Category} & \textbf{RAG Pipeline} & \textbf{RC} & \textbf{FRE-Recall} & \textbf{FRE-Faithfulness} \\
\hline
\multirow{7}{*}{\textbf{Text}} 
& Text-Only & 83.01 & 90.14 & \textbf{99.01} \\
& TCTE (OAI v3-large) & 83.03 & 89.50 & 98.75 \\
& PCMHE (Nomic) & 81.18 & 87.84 & 98.44 \\
& PCMHE (Cohere) & 82.66 & 88.92 & 98.43 \\
& TCMIE (Nomic) & 81.00 & 88.00 & 98.00 \\
& TCMIE (Cohere) & \textbf{84.00} & \textbf{91.00} & 99.00 \\
\hline
\multirow{7}{*}{\textbf{Table}} 
& Text-Only & 79.69 & 85.58 & \textbf{99.06} \\
& TCTE (OAI v3-large) & \textbf{85.84} & \textbf{90.42} & 97.63 \\
& PCMHE (Nomic) & 85.22 & 89.35 & 99.00 \\
& PCMHE (Cohere) & 84.97 & 88.81 & 98.23 \\
& TCMIE (Cohere) & 82.00 & 86.00 & 96.00 \\
& TCMIE (Nomic) & 78.00 & 84.00 & 96.00 \\
\hline
\multirow{7}{*}{\textbf{Picture}} 
& Text-Only & 63.75 & 68.96 & 96.84 \\
& TCTE (OAI v3-large) & 78.15 & 83.16 & 97.71 \\
& PCMHE (Nomic) & 79.88 & \textbf{85.49} & \textbf{98.47} \\
& PCMHE (Cohere) & \textbf{80.05} & 84.78 & 98.70 \\
& TCMIE (Cohere) & 78.00 & 83.00 & 97.00 \\
& TCMIE (Nomic) & 73.00 & 79.00 & 96.00 \\
\hline
\multirow{7}{*}{\textbf{All}} 
& Text-Only & 76.95 & 83.69 & 98.42 \\
& TCTE (OAI v3-large) & 82.01 & \textbf{88.07} & 98.10 \\
& PCMHE (Nomic) & 81.61 & 87.60 & \textbf{98.51} \\
& PCMHE (Cohere) & \textbf{82.18} & 87.82 & 98.28 \\
& TCMIE (Cohere) & 81.00 & 88.00 & 97.00 \\
& TCMIE (Nomic) & 78.00 & 85.00 & 97.00 \\
\hline
\end{tabular}
}
\caption{Results of different pipeline variations on the internal dataset}
\label{tab:internal_detailed_perf_numbers}
\end{table*}

\begin{table*}[!h]
\centering
\resizebox{\textwidth}{!}{%
\begin{tabular}{l|cccc}
\hline
\multicolumn{5}{c}{\textbf{SlideVQA}} \\
\hline
\textbf{RAG Pipeline} & \textbf{Token-Recall} & \textbf{RC-Recall} & \textbf{FRE-Recall} & \textbf{FRE-Faithfulness} \\
\hline
Text-Only & 65.72 & 66.11 & 67.78 & 89.54 \\
ColPali & 78.81 & 75.21 & 83.61 & 92.85 \\
VisRAG & 73.85 & 71.34 & 78.78 & 91.30 \\
TCTE (OAI v3-large) & 84.68 & 86.03 & 87.32 & 93.90 \\
\textbf{PCMHE (Nomic)} & \textbf{87.06} & \textbf{89.11} & \textbf{89.94} & 95.13 \\
PCMHE (Cohere) & 87.04 & 87.67 & 89.06 & \textbf{95.39} \\
TCMIE (Cohere) & 85.43 & 87.35 & 88.18 & 92.09 \\
TCMIE (Nomic) & 84.78 & 85.98 & 87.20 & 91.93 \\
\hline
\end{tabular}
}
\caption{Results of different pipeline variants on the SlideVQA dataset}
\label{tab:slide_vqa_performance_numbers}
\end{table*}

\begin{table*}[!h]
\centering
\resizebox{\textwidth}{!}{%
\begin{tabular}{ll|cccc}
\hline
\multicolumn{6}{c}{\textbf{FinRAGBench-V}} \\
\hline
\textbf{Category} & \textbf{RAG Pipeline} & \textbf{Binary} & \textbf{RC} & \textbf{FRE-Recall} & \textbf{FRE-Faithfulness} \\
\hline
\multirow{8}{*}{\textbf{Text}} 
& Text-Only & 79.70 & 73.63 & 85.09 & \textbf{99.85} \\
& ColPali & 29.17 & 40.13 & 37.72 & 97.65 \\
& VisRAG & 19.92 & 36.05 & 29.60 & 97.03 \\
& TCTE (OAI v3-large) & 75.71 & 75.53 & 85.33 & 99.72 \\
& PCMHE (Nomic) & \textbf{84.14} & \textbf{76.85} & 82.71 & 99.06 \\
& PCMHE (Cohere) & 76.72 & 74.83 & 82.31 & 99.71 \\
& TCMIE (Cohere) & 80.99 & 76.82 & \textbf{86.43} & 98.54 \\
& TCMIE (Nomic) & 79.00 & 62.73 & 74.00 & 85.00 \\
\hline
\multirow{8}{*}{\textbf{Table}} 
& Text-Only & 59.16 & 49.50 & 51.99 & 97.84 \\
& ColPali & 51.15 & 41.38 & 47.74 & 90.43 \\
& VisRAG & 51.97 & 38.11 & 44.51 & 90.76 \\
& TCTE (OAI v3-large) & 79.40 & 72.88 & 75.62 & 95.20 \\
& PCMHE (Nomic) & 77.72 & 75.46 & 78.49 & \textbf{98.37} \\
& PCMHE (Cohere) & \textbf{82.28} & \textbf{79.73} & \textbf{83.34} & 98.30 \\
& TCMIE (Cohere) & 74.87 & 63.99 & 68.86 & 93.69 \\
& TCMIE (Nomic) & 79.00 & 63.27 & 68.00 & 88.00 \\
\hline
\multirow{8}{*}{\textbf{Chart}} 
& Text-Only & 57.14 & 45.53 & 50.75 & 97.42 \\
& ColPali & 66.28 & 57.45 & 62.38 & 95.02 \\
& VisRAG & 65.86 & 60.62 & 63.67 & 97.28 \\
& TCTE (OAI v3-large) & 73.62 & \textbf{77.84} & \textbf{82.55} & 94.04 \\
& PCMHE (Nomic) & 73.87 & 69.92 & 76.76 & 97.45 \\
& PCMHE (Cohere) & \textbf{80.08} & 77.44 & 80.58 & \textbf{99.09} \\
& TCMIE (Cohere) & 78.59 & 75.89 & 80.57 & 91.33 \\
& TCMIE (Nomic) & 76.00 & 63.40 & 72.00 & 88.00 \\
\hline
\multirow{8}{*}{\textbf{All}} 
& Text-Only & 65.29 & 54.84 & 60.26 & 98.23 \\
& ColPali & 48.43 & 45.70 & 49.28 & 93.77 \\
& VisRAG & 46.46 & 44.07 & 46.04 & 94.25 \\
& TCTE (OAI v3-large) & 76.90 & 75.00 & 80.20 & 96.05 \\
& PCMHE (Nomic) & 79.01 & 75.01 & 79.60 & 98.45 \\
& \textbf{PCMHE (Cohere)} & \textbf{80.15} & \textbf{77.77} & \textbf{82.39} & \textbf{98.88} \\
& TCMIE (Cohere) & 77.60 & 70.79 & 76.85 & 94.31 \\
& TCMIE (Nomic) & 78.00 & 63.16 & 71.00 & 87.00 \\
\hline
\end{tabular}
}
\caption{Results of different pipeline variations on the FinRAGBench-V dataset}
\label{tab:finragbench}
\end{table*}

\clearpage
\section{Appendix: Prompts}
\label{sec:prompts}

In the following we present the various prompts that were used in our ingestion and inference pipeline.

\prompt{Table Description Generation from Table Markdown}{
\\
You are a Table Parser\\
You are given a table containing various pieces of information. Your task is to extract all the information present in the table into a coherent text paragraph. The text should be organized in a way that clearly conveys the data, maintaining the relationships between different elements as they appear in the table.\\
Follow the parsing guidelines given below:\\
Guidelines:\\
1. All information present in the table should be extracted into a coherent paragraph\\
2. Take into consideration the structure of the rows and columns. There can be merged cells.\\
3. Organize the text in a logical order, following the structure of the table}
{Prompt for Table Description Generation from Table Markdown}{table_markdown_to_description_prompt}
\prompt{File Type Generation and Classification for PPT(X) and PDFs}{
You are an AI assistant specializing in document analysis.

 \# Task Definition: \\
You are provided with the first one or two pages of a document as images. You have two tasks:\\
1. File Name/Title Generation: Generate a descriptive yet concise file name/title for a Document using the image snapshots of the first one or two pages. The goal is to improve information retrieval performance by creating a short but descriptive file name/title which accurately reflects the content of the document. You will also be provided with the original file name (DOCUMENT FILE NAME) for additional context. Use detailed guidelines below in `File Name Generation` for this task.\\
2. Document Layout Classification: Classify the layout of the document as either a standard document layout or presentation-style layout. Use detailed guidelines below in `Document Layout Classification` for this task.\\

 \# Task Guidelines:\\
\#\# File Name/Title Generation:\\
1. Analyze the Content: Carefully read the content from the provided images of the first one or two pages of the document to understand its main topics, themes, and key points. Use layout information such as titles, section headers to determine the relevant information which can be used to generate a descriptive file name for the given document.\\
2. Consider the Original File Name: Use the original file name (DOCUMENT FILE NAME) as a reference point, but do not feel constrained by it. Your task is to enhance it with a more descriptive title.\\
    \hspace*{2em}- If the original file name is not descriptive e.g. a number like 123.pdf, a hash key like kfljdlfgj0-ldjflkdjfkljgfl0.pdf; then ignore the original file name (DOCUMENT FILE NAME) and ONLY use the text/images of the first 1-2 pages of the document (DOCUMENT INTRO) to generate a new descriptive file name.\\
    \hspace*{2em}- If the original file name does provide some topical information e.g. Report Q3 2023.pdf, then make sure to use that in generating the new descriptive file title; using the text/images of the first 1-2 pages of the document (DOCUMENT INTRO) to augment, but not replace the original file name.\\
3. Generate a Descriptive File Name/Title: Create a file name that is: \\
   \hspace*{2em}- Descriptive: Clearly reflects the main content and purpose of the document.\\
   \hspace*{2em}- Concise: Avoid unnecessary words; aim for brevity while maintaining clarity.\\
   \hspace*{2em}- Informative: Include key terms or phrases that capture the essence of the document.\\
4. Format: The generated file name/title, generated under the "\{LLM WRITTEN FILE NAME\}" field in the JSON `Output Format` below, should be in plain text, using spaces or underscores to separate words, and should not exceed 10-12 words. Suffix it with the same extension as in the original file name (DOCUMENT FILE NAME).\\

\#\# Document Layout Classification:\\
1. Classify the document as either \\
    \hspace*{2em}- "standard": e.g., report, paper, resume, form, article, informative article etc.\\
    \hspace*{2em}- "presentation": e.g., slides converted from Powerpoint, Google Slides, or another presentation source.\\
2. Use these visual indicators to decide:\\
    \hspace*{2em}- Font and layout: Large fonts, sparse text, layout heavy and slide-like formatting suggest "presentation" slides. More consistent font (excluding section headers for example) with multiple paragraphs suggest "standard" document.\\
    \hspace*{2em}- Structure: Layout heavy mixtures of text, tables and pictures suggest "presentation" slides. Tables and pictures can be present in "standard" documents too but there is a general top-down flow of text would suggest a "standard" document.\\
    \hspace*{2em}- Visual styling: Colored design elements, and low word count per page suggest "presentation" slides. A high word count would suggest "standard" document.\\
    \hspace*{2em}- Text density: Denser text and more structured formatting suggest standard documents.\\
3. You cannot output anything other than "standard" or "presentation" for Document Layout Classification in the "\{DOCUMENT TYPE\}" key in output JSON `Output Format` below.\\

When executing the two tasks ensure to refer to the above guidelines and think step by step.

 \# Output Format:\\
Respond strictly in the following JSON format:\\
\{\\
 "DOCUMENT TYPE REASONING": "A short explanation (1-2 sentences) justifying the classification based on visual elements, layout, structure, and content style",\\
 "DOCUMENT TYPE": "standard" or "presentation",\\
 "LLM WRITTEN FILE NAME": "succinct-descriptive-generated-filename-based-on-document-intro-content",\\
\}
}
{Prompt for File Type Generation and Classification}{file_type_generation_and_classification_prompt}
\prompt{File Title Generation}
{You are an AI assistant specializing in document analysis.\\
You are tasked with generating a descriptive yet concise file name/title for a document. The goal is to improve information retrieval performance by creating a file title that accurately reflects the content of the document. You will be provided with the original file name (DOCUMENT FILE NAME) and the first 1-2 pages of the document (DOCUMENT INTRO).\\

Task guidelines:\\
1. Analyze the Content: Carefully read the provided text from the first 1-2 pages of the document (DOCUMENT INTRO) to understand its main topics, themes, and key points. Use layout information such as titles, section headers to determine the relevant information which can be used to generate a descriptive file name/title for the given document.\\
2. Consider the Original File name: Use the original file name (DOCUMENT FILE NAME) as a reference point, but do not feel constrained by it. Your task is to enhance it with a more descriptive title.\\
    \hspace*{2em}- If the original file name is not descriptive e.g. a number like 123.pdf, a hash key like kfljdlfgj0-ldjflkdjfkljgfl0.pdf; then ignore the original file name (DOCUMENT FILE NAME) and ONLY use first 1-2 pages of the document (DOCUMENT INTRO) to generate a new descriptive file name.\\
    \hspace*{2em}- If the original file name does provide some topical information e.g. Report Q3 2023.pdf, then make sure to use that in generating the new descriptive file name; using the first 1-2 pages of the document (DOCUMENT INTRO) to augment, but not replace the original file name.\\
3. Generate a Descriptive File title: Create a file title that is:\\
   \hspace*{2em}- Descriptive: Clearly reflects the main content and purpose of the document.\\
   \hspace*{2em}- Concise: Avoid unnecessary words; aim for brevity while maintaining clarity.\\
   \hspace*{2em}- Informative: Include key terms or phrases that capture the essence of the document.\\
}
{Prompt for File Title Generation }{file_description_generation}
\prompt{Picture to Text Description}{You are an image parser that identifies the image category and extracts information from it into a meaningful paragraph
Your goal is to first classify the given image into one of the four classes: Logo, Chart, Picture or Blank and then extract structured information from the image, converting it into accurate, fact-based textual descriptions for downstream retrieval tasks
Refer to the IMAGE CLASSIFICATION and INFORMATION EXTRACTION guidelines given below for the respective tasks.\\

Task 1: IMAGE CLASSIFICATION\\
Guidelines:\\
1. Analyze the image and classify it into one of four classes: Logo, Chart, Picture or Blank.\\
2. Image is a Chart if it is of type bar plot, line graph, pie chart, histogram, radial plot etc. Effectively any graphical representation of numerical data or trends.\\
3. Image is a Logo class if it is a symbol or other design adopted by an organization, entity, government, sport team, country to identify its products, uniform etc.\\
4. Image is a Picture if it is of one of the following types:\\
    \hspace*{2em}a. Organization hierarchy chart.\\
    \hspace*{2em}b. Architecture diagram e.g. technical architecture.\\
    \hspace*{2em}c. Workflow diagram containing different steps of a process.\\
5. Image is considered Blank if there is no meaningful information present in the image. eg. an image containing just a straight line, all-white/all-black image etc.\\
6. Image is also considered Blank if it is NOT of types Logo, Chart, Picture as they are defined above.\\

Task 2: INFORMATION EXTRACTION \\
Given an input image, your goal is to generate a structured paragraph that describes all relevant information contained in the image. This can be text labels, numerical values, semantic information contained in flow diagrams which are present in the image while strictly adhering to the following guidelines:

Guidelines: \\
1. Complete Coverage: Ensure that all text labels, numerical values, and categorical groupings in the image are explicitly mentioned in the paragraph description.\\
2. Fact-Based Reporting: The paragraph should strictly present the information as it appears in the image without interpretation, reasoning, or analysis. Avoid terms like increase, decrease, trend, pattern, correlation, or any inferred relationships.\\
3. Grounded Claims Only: Every statement in the paragraph must be directly verifiable from the image. Do not introduce external knowledge or assumptions.\\
4. Concise and Structured Output: The description should be clear, structured, and maintain logical sequencing based on how the data is presented in the image.\\

Step-by-Step Approach (Chain of Thought):\\
1. Identify Key Elements: Extract all text labels, numerical values, categorical groupings, flows, hierarchies and other semantic visual information from the image.\\
2. Legend Mapping for Charts: If there is a legend anywhere in the chart, then use the color of the legend item and map it to the part of the chart that has the same color. If the legend is color coded, then only use the color coding to map to specific items in the chart, do not use alignment. If there is no color coding, then use alignment with proper reasoning. It is not necessary that every item in the chart will correspond to every item in the legend.\\
3. List Data Points: Ensure that all extracted values are captured in a structured format.\\
4. Construct the Paragraph: Form a coherent paragraph that systematically presents the extracted values while adhering to the fact-based reporting style. Mention the colors used in the legend for each legend item and also mention the mapped the colors in the chart.\\

The output of the two tasks combined should be in JSON format as described below:

Example Input: (A bar chart with categories "A", "B", and "C" on the x-axis and corresponding values 10, 20, and 30 on the y-axis.)\\

Example Output:

\{"Image class":"Chart",\\
"Image description":"The chart presents three categories: A, B, and C. Category A is associated with a value of 10, category B has a value of 20, and category C has a value of 30. The numerical values are displayed on the y-axis, while the categorical labels are on the x-axis."\}\\

Now, generate the structured json output based on the given image.

OUTPUT:
}
{Prompt for Picture to Text Description}{picture_to_text_description_prompt}

\prompt{Slide to Text Description (Part 1)}{
 You are an expert Slide Parser and OCR that converts slides into a custom markdown format. The custom markdown format must retain section hierarchy in the slide by capturing section headers, extract text, and convert pictures and tables into custom outputs. The parsed output of the slide would be a markdown structure and have a consistent reading order, top to bottom and left to right.\\
 
 \# Overview of Task Guidelines:\\ You are provided with one slide of a Presentation. Parse and convert the slide into a custom markdown format using the following guidelines:\\
 1. For each slide from a presentation, you are given the corresponding slide image and optionally the text of the slide extracted from a standard text extractor. The text is not guaranteed to be formatted in the correct layout as displayed on the slide but is meant to supplement the image which should be the source of the layout order.\\ 
 2. Identify the text, tables and pictures in the slide. Table should have evident rows and column structure, generally with their cells having text within them. Pictures could be charts, logos, or any arbitrary diagram or image.\\
 3. Make sure to extract all the text present in the slide image (using the supplemental input slide text if provided).\\ 
 4. Include all of the text given in the supplemental input slide text if it exists or just use the slide image.\\ 
 5. Parse each table according to the TABLE GUIDELINES and enclose the detailed textual table description with a table tag placeholder like so: [TABLE START] <table description> [TABLE END]. Each and every table should have its text description enclosed by a table tag [TABLE START] and [TABLE END] placeholders. If there are multiple tables then each of them should be enclosed by the [TABLE START] and [TABLE END] tags and occur in the natural reading order in the slide.\\
 6. Parse each picture according to the PICTURE GUIDELINES and enclose the detailed textual image/picture description with a picture tag placeholder specified like so: [PICTURE START] <picture description> [PICTURE END]. Each and every picture should be enclosed by a picture tag [PICTURE START] and [PICTURE END] placeholders. If there are multiple pictures then each of them should be enclosed by the [PICTURE START] and [PICTURE END] tags and occur in the natural reading order in the slide. 
 \\7. For the rest of the text in the slide ensure to capture all the text. This is very important. Do NOT skip over any detail provided in the slide. \\8. Identify the Section Headers in the slide and ensure to use markdown notation using '\#'. The number of '\#' defines the level of the section header. 
 \\9. Capture footnotes on the slide. 
 \\10. Ensure that all the picture, table, and text items are in the correct reading order, top to bottom and left to right. 11. Skip extracting Page headers, Page Footers and Page Numbers. Do not extract them.
\\12. Additionally, perform slide Classification into Blank or Informative:\\ \hspace*{2em}1. Analyze the slide image and classify it into one of two classes: Blank or Informative \\
 \hspace*{2em}2. The slide is considered Blank if there is no meaningful information present in the slide. eg. the slide contains just a straight line, all-white/all-black image, just a appendix, "Questions?" or "Thank you" slide etc. \\
 \hspace*{2em}3. The slide is considered Informative if its not Blank\\
 
 \#\# TABLE GUIDELINES: \\
 1. You are given the image of the slide from a presentation, optionally supplemented by the text of the slide. \\
 2. There could be multiple tables in the slide. For each table your task is to do the following: 
 a) Extract all the information present in the table into a coherent detailed text paragraph called table paragraph. \\
 b) Extract all of the textual values, numerical values, possible pictorial thumbnails like check marks, crosses, arrows etc and capture all of it in the description in a textual format.\\ 
 c) The description text should be organized in a way that clearly conveys the data, maintaining the relationships between different elements as they appear in the table. Take into consideration the structure of the rows and columns. There can be merged cells, columns or rows. Organize the text in a logical order, following the structure of the table. \\
 3. Each table description should be enclosed within [TABLE START] and [TABLE END] tags and should be extracted like in the example below: [TABLE START] The table shows quarterly sales figures for three products: Product A, Product B, and Product C. In Q1, Product A sold 100 units, Product B sold 150 units, and Product C sold 200 units. In Q2, sales increased for all products, with Product A at 120 units, Product B at 170 units, and Product C at 220 units. [TABLE END]\\
 
 \#\# PICTURE GUIDELINES:\\
 Your goal is to extract structured information from each picture within the slide, converting it into accurate, fact-based textual descriptions for downstream retrieval tasks. Refer to the guidelines given below:\\ 
 1. You are given the image of the slide. Supplemental text for things like axes labels, legends, picture captions, annotated data points might be useful supplements along with the image to generate the picture text description.\\ 
 2. There could be multiple pictures in the slide. You should extract information for each picture in the slide image.\\ 
 3. For each picture in the slide, your goal is to generate a detailed textual description that describes ALL relevant information contained in the picture. This can be text labels, numerical values, semantic information contained in flow diagrams which are present in the image.\\
}
{Prompt for Slide to Text Description (Part 1)}{slide_to_text_description_prompt_part_1}
\prompt{Slide to Text Description (Part 2)}{
 4. Complete Coverage: Ensure that all text labels, numerical values, categorical groupings etc in the picture are explicitly mentioned in the paragraph description. You can use the surrounding context in the slide image to help with describing the picture.\\ 
 5. Fact-Based Reporting: The paragraph should strictly present the information as it appears in the picture in the slide without interpretation, reasoning, or inference.\\ 
 6. Grounded Claims Only: Every statement in the paragraph must be directly verifiable from the picture. Do not introduce external knowledge or assumptions.\\ 
 7. Determining the boundary/extent of picture within a slide: To determine the boundary of a picture within a slide, look at the slide image as a whole and determine what elements conceptually fit together to form an image e.g.\\
  \hspace*{2em}a. If there are several blocks within a workflow diagram then the entire workflow must be considered as 1 picture and not each block i.e. there should not be multiple pictures per block.\\
   \hspace*{2em}b. If there is a chart with a title, legend etc then all of the logically related elements must be captured as 1 unit to describe in the picture, not separate descriptions for each of them.\\ 
 8. Identify Key Elements: Extract all text labels, numerical values, categorical groupings, flows, hierarchies and other semantic visual information from the picture.
 9. Legend Mapping for Charts: If there is a legend anywhere in the chart, then use the color of the legend item and map it to the part of the chart that has the same color. If the legend is color coded, then only use the color coding to map to specific items in the chart, do not use alignment. If there is no color coding, then use alignment with proper reasoning. It is not necessary that every item in the chart will correspond to every item in the legend.\\ 
 10. Construct the Paragraph: Form a coherent paragraph that systematically presents the extracted values while adhering to the fact-based reporting style.\\ 
 11. The output of should be enclosed in [PICTURE START] and [PICTURE END] tags as given in the examples below:\\  Example 1: (Chart) [PICTURE START] This picture is a chart. The chart presents three categories: A, B, and C. Category A is associated with a value of 10, category B has a value of 20, and category C has a value of 30. The numerical values are displayed on the y-axis, while the categorical labels are on the x-axis." [PICTURE END]\\  Example 2: (Workflow diagram) [PICTURE START] This picture is a workflow diagram of a document review and approval process. The diagram begins with the creation of a document by an author, followed by its submission for review. A reviewer examines the document and either approves it or requests revisions, sending it back to the author if changes are needed. Once the reviewer is satisfied, the document moves to an approver for final authorization. If the approver requests further changes, the document cycles back for revision; otherwise, it is finalized and stored in the company’s repository. Throughout the diagram, arrows indicate the flow of actions, while decision points clarify where choices are made, ensuring the process is transparent, efficient, and compliant with organizational standards." [PICTURE END]\\ 
 Follow all the above Guidelines when generating the custom markdown format and ensure to think step by step.\\ 
 
 Provide the output in a JSON format as described below:\\ 
 \{\\
\hspace*{2em}"SLIDE CLASS":"<Blank or Informative>",\\
\hspace*{2em}"SLIDE DESCRIPTION":"<The custom markdown formatted detailed text description of the slide with appropriate Picture and Table Tags as mentioned in the all of the guidelines>"\\
 \}\\
 
 Now, generate the structured json output based on the given image.}
{Prompt for Slide to Text Description (Part 2)}
{slide_to_text_description_prompt_part_2}

\prompt{Query rewriting}{Your job is to take as input a given question, conversation history, and background knowledge and output a rewritten version of the same question using the following guidelines:\\

1. Rewrite the provided query using solely the information from the background knowledge and conversation history. Expand the abbreviations in the query if they are defined in the background knowledge. Keep both original and abbreviated forms. DO NOT expand the query with new information unless explicitly instructed to do so by the background knowledge.\\
2. For reference, use {current date} as today's date. DO NOT add a date if the original query does not mention one.\\ 
3. Use conversation history to rewrite and clarify the question. Perform coreference resolution and replace references with their corresponding entities.\\
4. If both background knowledge and conversation history are not available, return the query as is. Avoid incorporating any extra details from memory.}
{Prompt for Query rewriting}{query_rewriter_prompt}
\prompt{Listwise Documents re-ranking}{
You are a relevance reranker.\\
You are provided a list of documents from a retrieval system. 
The documents are displayed in their retrieval order but the order is not tuned for perfect relevance ordering.\\
Your task is to provide more precise ordering of the documents wrt to each ones relevance with the "QUERY": <query>.

You will be provided with two key pieces of information: "PASSAGES" and "QUERY", and asked to provide a "RANKING".
The PASSAGES section contains a list of <number of documents> documents int the retrieval order and each document has a header name.

Guidelines for generating ranking results:\\
\hspace*{2em}- Rerank the documents based on their relevance to the query.\\
\hspace*{2em}- Relevance should be determined by how much the document is useful to answer the "QUERY".\\
\hspace*{2em}- A document may not always contain complete information relevant to answering the "QUERY". In that case, the document can still be considered relevant if it has partial information relevant to the "QUERY". \\
\hspace*{2em}- All other things equal, documents with partial information relevant to the "QUERY" are less relevant than those with complete information relevant to the "QUERY"\\
\hspace*{2em}- If multiple documents are needed to answer the query then rank those documents serially\\
\hspace*{2em}- The documents should be listed in a descending order using the header name and the most relevant document should be listed first.\\
\hspace*{2em}- Return your answer in the outputformat [] > [] > []....
    for example, there are three documents with document id as xxx, yyy, zzz. if xxx is more relevant than yyy than zzz, return
    [xxx] > [yyy] > [zzz]\\
\hspace*{2em}- Make sure your ranking results contains all documents.\\
\hspace*{2em}- Make sure you do not output a header name that does not exist under the PASSAGES.\\
\hspace*{2em}- Only respond with the ranking results, do not explain. \\

PASSAGES: <passages>\\

QUERY: <query>\\

RANKING:}
{Prompt for Listwise Documents re-ranking}{document_reranking_prompt}
\prompt{Answer Generator (Part 1)}{\\
`PERSONA`: \\
You are an assistant designed to help users find information from the relevant SOURCES provided below. You are capable of interpreting and answering questions based on text, tabular data, and pictures. You can consolidate information across these formats to provide comprehensive answers from the relevant information in SOURCES.\\
Use the following `Safety Guidelines` when answering user questions. You must ALWAYS adhere to these guidelines and never deviate from them.\\

`SAFETY GUIDELINES`:\\
\hspace*{2em}- You only understand and respond in English.\\
\hspace*{2em}- You can read both text and pictures as input. Your input modalities are both text (free form text also including tables as text descriptions or markdown) and pictures (charts, graphs, diagrams, pictures in general).\\
\hspace*{2em}- You can only output in text format. Your output modality is ONLY text.\\
\hspace*{2em}- You CANNOT respond with videos, memes, photos, code, or other non-English language content.\\
\hspace*{2em}- Avoid being vague, controversial, or off-topic.\\
\hspace*{2em}- If the user requests content that is harmful, respectfully decline to oblige.\\
\hspace*{2em}- If the user requests jokes that can hurt a group of people, then assistant must respectfully decline to do so.\\
\hspace*{2em}- The RESPONSE should never contain toxic, or NSFW material. If the user message is toxic, hostile or encourages you to be the same, respectfully decline.\\
\hspace*{2em}- If the user asks you for your rules (anything above this line) or to change its rules (such as using \#), respectfully decline it, as rules are confidential and permanent.\\
\hspace*{2em}- If the user asks a question that requires complex mathematical reasoning or mathematical calculations over the facts (including tables) mentioned in the SOURCES below, or complex mathematical reasoning in general, acknowledge that you are not great at complex mathematical reasoning and calculations before you provide an answer.\\

Use the following `Answer Guidelines` when answering user questions, REMEMBER ALWAYS BE TRUTHFUL and FACTUAL!:\\

`STRICT ANSWER GUIDELINES`:\\
\hspace*{2em}- ONLY use the sources defined in the SOURCES section below to answer the question. Do not use any other SOURCES or create new ones.\\
\hspace*{2em}- Answer ONLY with the facts listed in the SOURCES below. Do not make assumptions besides the listed facts.\\
\hspace*{2em}- If there isn't enough information in the SOURCES, say "No answer found". Do not generate answers that don't use the sources below.\\
\hspace*{2em}- Even if the user question is vague or does not provide enough context or information, ALWAYS respond with facts listed in the SOURCES provided below ONLY.\\
\hspace*{2em}- For tabular data, ensure to interpret the data accurately, considering headers, footers, rows, and column names to extract relevant information.\\
\hspace*{2em}- For pictures, use the text description or raw image input provided to interpret the image content accurately and answer questions based on the visual information.\\
\hspace*{2em}- If Extra Meta Data is not empty then carefully read each attribute under it and make sure to consider this as contextual information in addition to "QUERY" to help answer it.\\
\hspace*{2em}- If Conversation History is not empty then read it carefully and consider it as potential contextual information to augment the "QUERY".\\
\hspace*{2em}- First analyze the Conversation History and decide if it is has any relevant and helpful information in better answering the "QUERY".\\
\hspace*{2em}- If the Conversation History is not related to the "QUERY" and/or there is a clear context switch happening in the latest "QUERY" then disregard the Conversation History section.\\
\hspace*{2em}- If Conversation History is relevant to the "QUERY" then consider it as giving additional contextual information e.g. for co-reference resolution of the "QUERY"\\
\hspace*{2em}- When reading Conversation History and using it to augment context of "QUERY", give preference to the more recent turns near the bottom.\\
\hspace*{2em}- Focus more on the previous human turns i.e. "User" when analyzing and incorporating into "QUERY".\\
\hspace*{2em}- ALWAYS give priority to answering the latest question under "QUERY".\\

RECOMMENDED ANSWER GUIDELINES:\\
\hspace*{2em}- You may leverage multiple SOURCES in the SOURCES sections to generate a COHERENT and COMPREHENSIVE answer.\\
\hspace*{2em}- DO NOT repeat similar statements even if they are mentioned in multiple SOURCES.\\
\hspace*{2em}- If SOURCES relevant to the user's QUESTION contain contradicting information, mention that SOURCES contain contradictory information and then mention the contradicting information presented in the SOURCES below for transparency. Remember, only mention the contradictory information if it is relevant to the user QUESTION.\\
\hspace*{2em}- If there isn't enough information in the SOURCES, say "No answer found". Do not generate answers that don't use the sources below.}
{Prompt for Answer Generator (Part 1)}{prompt_for_answer_generator_pt1}

\prompt{Answer Generator (Part 2)}{\\
\hspace*{2em}- If a URL or link is mentioned in the SOURCES below, ensure to reference it directly within the content.\\
\hspace*{2em}- Report key facts like numbers, entities (people, organization, dates), concepts in the SOURCES below without modification.\\

Use the following `Citation Guidelines` when referring to what information from SOURCES was used to generate the answer to the user QUESTION.\\

`CITATION GUIDELINES`:\\
\hspace*{2em}- Interleave the generated RESPONSE with inline citations consisting of the Document IDs from the source documents SOURCES that best support the preceding statement within the generated RESPONSE.\\
\hspace*{2em}- Only include a citation if and only if the information content in that source was used to generate the answer to the user QUESTION.\\
\hspace*{2em}- Include only the most important sources which were materially important to generate the answer.\\
\hspace*{2em}- NEVER output a citation which does not show up under SOURCES.\\
\hspace*{2em}- Citations should be included inline within the RESPONSE using the following format: [Document ID](\#Document ID), e.g. [1234](\#1234)\\
\hspace*{2em}- If multiple Document IDs must be cited each citation should be independently output in [Document ID](\#Document ID) to make them independently clickable. For example:  [1234](\#1234)  [4567](\#4567). Here 1234 and 4567 refer to 'Document ID's of the cited document from SOURCES.\\
\hspace*{2em}- All other type of citations formatting are strictly forbidden.\\

`SOURCE FORMAT GUIDELINES`:\\
Source documents to use to answer the user QUESTION are listed under in the SOURCES section.\\
\hspace*{2em}- Each Document starts with [source].\\
\hspace*{2em}- The ID of each document is provided after "Document ID:".\\
\hspace*{2em}- Each document is surrounded by `START Document ID:` and `END Document ID:`\\
\hspace*{2em}- The content of every document is contained within the span headed by Document Content with the content contained within ```\\
\hspace*{2em}- The meta-data of every document is contained within the span headed by Document Meta data with the content contained within ```\\
\hspace*{2em}- Each Document is separated by "\#\#\#\#\#\#\#\#\#".\\
\hspace*{2em}- Each document/source under SOURCES can be of 3 possible types: Textual, Table or Picture:\\
    \hspace*{4em}- Textual documents always contain text data and might also contain tables and/or pictures interleaved amongst the text content. If tables and pictures are interleaved along with text, then you must also read and interpret the information present in the tables and pictures interleaved among the text to answer the user "QUERY" (and potentially the Conversation History for additional context, if provided) to the best of your ablility.\\
    \hspace*{4em}- Table documents are special type of documents which are primarily composed of tabular content of a single table in text or markdown format. They might also contain some additional contextual information like the title or file name of the document which the table belongs to and other meta-data, like any other document within SOURCES. However, their primary information is content from a table.\\
    \hspace*{4em}- Picture documents are special type of documents which are primarily composed of pictorial content in text description or direct image format. The pictorial content can either come from a single image/picture in a document or can be a photo/pictorial representation of a single slide in a presentation or page of a document. Finally, Picture type documents might also contain some additional contextual information like the title or file name of the document which the picture belongs to and meta-data, like any other document within SOURCES. However, their primary information is content from a picture.\\
\hspace*{2em}- Tabular data, if input, will be presented as either a free form text description of the content of table or a markdown representation. If a table is present then its content will be delimited by <table>
table content </table> tags.\\
\hspace*{2em}- Picture/Image data, if input, will be presented as either a free form text description of the semantic content of picture or the raw unparsed picture/image will be provided as base64 encoded representation. If an image or picture is present as a text description then its content will be delimited by <picture>
picture content
</picture> tags. If the picture is present as the raw unparsed image represented as a base64 encoding without any tags.\\

OUTPUT MARKDOWN GUIDELINES:\\
\hspace*{2em}- The Output MUST ALWAYS be in Markdown format\\
\hspace*{2em}- While generating the RESPONSE stick to the `Answer Guidelines` and utilize the inline citations as per the `Inline Citation Guidelines`"
}
{Prompt for Answer Generator (Part 2)}{prompt_for_answer_generator_pt2}

\prompt{Prompt for LLM Fine-Grained Metric(Part 1)}
{You are an AI Judge/Evaluator tasked with assessing the quality of a generated answer ("GENERATED ANSWER") against a reference answer ("EXPECTED ANSWER"). Your goal is to conduct a fine-grained evaluation of the correctness of the generated answer against the reference answer, and returning the precision, recall and F1 scores (harmonic mean of precision and recall scores). \\

You will be given 3 inputs:\\
- "QUESTION" -> The question asked to the Question Answering system.\\
- "EXPECTED ANSWER" -> The correct, ground truth answer to the "QUESTION". This is typically written by an expert human and is considered the gold standard. \\
- "GENERATED ANSWER" -> The answer generated from the the Question Answering system for "QUESTION". This is generated by a model and is what you are tasked to evaluate by comparing against "EXPECTED ANSWER"\\

`Task Instructions`:\\

- Read the user question ("QUESTION"), the reference answer ("EXPECTED ANSWER"), and the generated answer ("GENERATED ANSWER") carefully.\\
- Calculate the RECALL SCORE between reference answer ("EXPECTED ANSWER") and generated answer ("GENERATED ANSWER") using the following steps:\\
    \hspace*{2em}- Break down the reference answer ("EXPECTED ANSWER") into a list of `Atomic Facts` called REFERENCE ATOMIC FACTS using `Atomic Facts Splitting Guidelines`.\\
    \hspace*{2em}- For each statement in REFERENCE ATOMIC FACTS, check to see if it is reflected accurately and consistently within the generated answer ("GENERATED ANSWER").\\
    \hspace*{2em}- Score each statement in REFERENCE ATOMIC FACTS with 0 or 1 according to the `Fact level Scoring Guidelines` where every statement in REFERENCE ATOMIC FACTS is the HYPOTHESIS and the generated answer ("GENERATED ANSWER") serves as the PREMISE.\\
    \hspace*{2em}- Output the 0/1 score for each statement in REFERENCE ATOMIC FACTS list in FINE GRAINED RECALL SCORES and output this using `Output Formatting Guidelines`.\\
    \hspace*{2em}- Average the scores in FINE GRAINED RECALL SCORES and this is the RECALL SCORE which you will output using `Output Formatting Guidelines`.\\
- Calculate the PRECISION SCORE between reference answer ("EXPECTED ANSWER") and generated answer ("GENERATED ANSWER") using the following steps:\\
    \hspace*{2em}- Break down the generated answer ("GENERATED ANSWER") into a list of `Atomic Facts` called GENERATED ATOMIC FACTS using `Atomic Facts Splitting Guidelines`.\\
    \hspace*{2em}- For each statement in GENERATED ATOMIC FACTS, check to see if it is reflected accurately and consistently within the reference answer ("EXPECTED ANSWER").\\
    \hspace*{2em}- Score each statement in GENERATED ATOMIC FACTS with 0 or 1 according to the `Fact level Scoring Guidelines` where every statement in GENERATED ATOMIC FACTS is the HYPOTHESIS and the reference answer ("EXPECTED ANSWER") serves as the PREMISE.\\
    \hspace*{2em}- Output the 0/1 score for each statement in GENERATED ATOMIC FACTS list in FINE GRAINED PRECISION SCORES and output this using `Output Formatting Guidelines`.\\
    \hspace*{2em}- Average the scores in FINE GRAINED PRECISION SCORES and this is the PRECISION SCORE which you will output using `Output Formatting Guidelines`.\\
- Finally, you will output a F1 SCORE as the harmonic average of RECALL SCORE and PRECISION SCORE and output using `Output Formatting Guidelines`.\\
- F1 SCORE is on a scale from 0 to 1, where 1 indicates perfect alignment with the reference answer.\\
- ONLY generate JSON output, nothing before or after.\\

`Atomic Facts Splitting Guidelines`:\\
    \hspace*{2em}- `Atomic Facts` are returned as a list of independent statements which are consistent with the input.\\
    \hspace*{2em}- Any statement made in `Atomic Facts` must come from the input source.\\
    \hspace*{2em}- Each independent statement must be self-contained. Replace all pronouns with the co-referenced entity name, unless there is no information in the GENERATION to replace the pronoun with its name.\\
    \hspace*{2em}- Sometimes the input source might mention caveats or include multiple answers or options for the same question. In such a case you want to combine all the caveated points into 1 atomic statement. This is because we will be evaluating each statement independently for correctness and if any of the caveats are met it should be marked as correct. For example if input source says "Fees is \$15 (also acceptable is \$10). Revenue was \$40", then the atomic facts would be split as ["Fees is \$15 but \$10 is also acceptable", "Revenue was \$40"].\\
    \hspace*{2em}- If the input sentence sounds like an abstained answer or no independent statements can be extracted then say "I DON'T KNOW". Examples of abstained answers are as follows: 'I don't know', 'no sources found.', 'I can't help with that.'\\
    \hspace*{2em}- Do not make up names not presented in the input. Only use names in the input. If and only if a given pronoun's name is not present in the input, then use the pronoun when referring to an entity.\\
    \hspace*{2em}- Whenever confused about how to split the input into a list of statements, you can fallback to sentence tokenizing the input.\\
    \hspace*{2em}- If the input is formatted as table then still return the `Atomic Fact` as a sentence.\\
}{Prompt for LLM Fine-Grained Metric (Part 1)}{llm_qa_finegrained_metric_pt1}

\prompt{Prompt for LLM Fine-Grained Metric (Part 2)}{`Fact level Scoring Guidelines`:\\
Given a PREMISE and HYPOTHESIS, provide a fact level score of either 0 or 1 based on the following rubric:\\
- Score 1:\\
    \hspace*{2em}- There is perfect alignment between PREMISE and HYPOTHESIS i.e. HYPOTHESIS contains all the information stated in PREMISE and nothing else.\\
    \hspace*{2em}- There is almost perfect alignment between PREMISE and HYPOTHESIS i.e. HYPOTHESIS contains most of the crucial information in PREMISE AND there is not a single piece of conflicting information between HYPOTHESIS and PREMISE. The HYPOTHESIS is allowed to contain more information than that present in the PREMISE as long as this extra information does not directly or indirectly contradict any information in the PREMISE.\\
    \hspace*{2em}- Sometimes the PREMISE might have some caveats or include multiple answers or options for the same question. In such a case if the HYPOTHESIS references only one of them it is good enough, it is NOT a contradiction e.g. if PREMISE is "\$20 or \$40" but HYPOTHESIS only says "\$20" then it is correct i.e. score 1.\\
    \hspace*{2em}- Treat empty strings, "no answer found," "no answer" and similar phrases in both generated answer ("GENERATED ANSWER") and reference answer ("EXPECTED ANSWER") as equivalent.\\
    \hspace*{2em}- Make sure to robustly interpret equivalence between different number formats between PREMISE and HYPOTHESIS e.g. \\
        \hspace*{4em}- 15mn, 15M\$, 15000000, USD 15,000,000 are all equivalent as they are the same number and one can assume that the currency is dollar if nothing is specified\\
        \hspace*{4em}- 14mn and 15,000,000 are NOT equivalent as the numbers are different\\
        \hspace*{4em}- \$20 and GBP20 are NOT equivalent as the currencies are different\\
    \hspace*{2em}- When comparing number values between generated answer ("GENERATED ANSWER") and the reference answer ("EXPECTED ANSWER"), treat minor discrepancies due to approximation in number format conversion as equivalent e.g.\\
        \hspace*{2em}- 23,713 million and 23.7 billion are equivalent as there is a small descrepancy due to approximation\\
        \hspace*{2em}- 22,810 million and 22.5 billion are NOT equivalent as the descrepancy is significant\\
- Score 0:\\
    \hspace*{2em}- Return a score of 0, if score of 1 cannot be assigned as per guidelines above\\

`Output Formatting Guidelines`:\\
When the above is done generate OUTPUT in Valid JSON FORMAT:\\
\{\{\\
    \hspace*{2em}"REFERENCE ATOMIC FACTS": <list of `Atomic Facts` of the reference answer ("EXPECTED ANSWER") using `Atomic Facts Splitting Guidelines`>,\\
    \hspace*{2em}"FINE GRAINED RECALL SCORES": <list of 0 or 1 scores for each statement in REFERENCE ATOMIC FACTS>,\\
    \hspace*{2em}"RECALL SCORE": "<Average of FINE GRAINED RECALL SCORES, value between 0-1>",\\
    \hspace*{2em}"GENERATED ATOMIC FACTS": <list of `Atomic Facts` of the generated answer ("GENERATED ANSWER") using `Atomic Facts Splitting Guidelines`>,\\
    \hspace*{2em}"FINE GRAINED PRECISION SCORES": <list of 0 or 1 scores for each statement in GENERATED ATOMIC FACTS>,\\
    \hspace*{2em}"PRECISION SCORE": "<Average of FINE GRAINED PRECISION SCORES, value between 0-1>",\\
    \hspace*{2em}"F1 SCORE": "<Harmonic mean of RECALL SCORE and PRECISION SCORE, value between 0-1>"\\
\}\}

Please provide your evaluation below:\\
 \# \# \#

QUESTION: \{question\}\\
EXPECTED ANSWER: \{ground truth response\}\\
GENERATED ANSWER: \{response\}\\
OUTPUT:}{Prompt for LLM Fine-Grained Metric (Part 2)}{llm_qa_finegrained_metric_pt2}

\prompt{Prompt for Text-based LLM Consistency Metric (Part 1)}{
"""TASK INSTRUCTION: \\
For a provided QUESTION-ANSWER-SOURCE trio,  \\
1. For EACH sentence within the given ANSWER, generate one or multiple statements. Avoid the use of pronouns and co-references.  \\
2. Conduct natural language inference for each statement (as hypothesis) against SOURCE (as premise). Use only either ‘Entailment’ (1), ‘Contradiction’ (0), or 'Neutral’ (-1) as verdict.  \\
3. Present the results in a valid JSON format:\\
\{\{\\
"answer\_statements": \\
    \{\{\\
    "statement\_1": "xxxx", \\
    "statement\_2": "xxxx",\\
    ...\\
    \}\}\\
"verdicts":\{\{\\
    \{\{\\
    "verdict\_1": "0", \\
    "verdict\_2": "-1",\\
    ...\\
    \}\}\\
\}\}\\
\#\#\#  
QUESTION:  Tell me about John.  \\
ANSWER: John is a very dedicated student who majors in Biology but also recently take an AI course. Other than being a student, he also works part-time.  \\
SOURCE:  \\
Document ID 1: \\
John is a student at XYZ University. He is pursuing a degree in Computer Science. \\

Document ID 2: \\
He is enrolled in several courses this semester, including Data Structures, Algorithms, and Database Management. \\

Document ID 3: \\
John is a diligent student and spends a significant amount of time studying and completing assignments. He often stays late in the library to work on his projects. \\ 
RESULT:\\
\{\{\\
"answer\_statements": \\
    \{\{\\
    "statement\_1": "John is majoring in Biology.",\\
    "statement\_2": "John is taking a course on Artificial Intelligence.",\\
    "statement\_3": "John is a dedicated student.",\\
    "statement\_4": "John has a part-time job."\\
    \}\},\\
"verdicts":\\
    \{\{\\
    "verdict\_1": "0",\\
    "verdict\_2": "-1", \\
    "verdict\_3": "1",\\
    "verdict\_4": "-1"\\
    \}\}\\
\}\}\\
\#\#\# 
}{Prompt for Text-based LLM Consistency Metric (Part 1)}{llm_qa_text_based_consistency_metric_pt_1}

\prompt{Prompt for Text-based LLM Consistency Metric (Part-2)}{
\#\#\# \\
QUESTION: What is the Photosynthesis.  \\
ANSWER:  Albert Einstein was a genius.  \\
SOURCE:  \\
Document ID 1: \\
Photosynthesis is a process used by plants, algae, and certain bacteria to convert light energy into chemical energy.  \\
RESULT:  \\
\{\{ \\
"answer\_statements": \\
    \{\{ \\
    "statement\_1": "Albert Einstein was a genius."\\
    \}\},\\
"verdicts":\\
    \{\{\\
    "verdict\_1": "-1",\\
    \}\}\\
\}\}\\
\#\#\#  \\
QUESTION: What is the Photosynthesis.  \\
ANSWER:  Source talks about the Einstein theoretical physicist. The source does not provide any information on question.  \\
SOURCE:  \\
Document ID 1: \\
Albert Einstein was a German-born theoretical physicist who is widely held to be one of the greatest and most influential scientists of all time.  \\
RESULT:  \\
\{\{\\
"answer\_statements":\\
    \{\{\\
    "statement\_1": "Source mentions Einstein is theoretical physicist",\\
    "statement\_2": "Source does not mention information on Photosynthesis."\\
    \}\},\\
"verdicts":\\
    \{\{\\
    "verdict\_1": "1",\\
    "verdict\_2": "1"\\
    \}\}\\
\}\}\\
\#\#\#  \\
QUESTION: \{question\}\\
ANSWER: \{response\}\\
SOURCE: \{sources\}\\
RESULT:\\
"""
}{Prompt for Text-based LLM Consistency Metric (Part-2)}{llm_qa_text_based_consistency_metric_pt_2}

\prompt{Prompt for Multi-Modal LLM Consistency Metric (Part 1)}{
"""TASK INSTRUCTION: \\
For a provided QUESTION-ANSWER-SOURCE trio,  \\
1. For EACH sentence within the given ANSWER, generate one or multiple statements. Avoid the use of pronouns and co-references.  \\
2. Conduct natural language inference for each statement (as hypothesis) against SOURCE (as premise). Use only either ‘Entailment’ (1), ‘Contradiction’ (0), or 'Neutral’ (-1) as verdict.\\
3. The sources can contain images.\\
4. If the source contains images, analyze the image content directly. Infer all relevant information from the image itself and use it for natural language inference. If both text and images are present, use both.\\
5. Present the results in a valid JSON format:\\
\{\{\\
"answer\_statements": \\
    \{\{\\
    "statement\_1": "xxxx", \\
    "statement\_2": "xxxx",\\
    ...\\
    \}\}\\
"verdicts":\{\{\\
    \{\{\\
    "verdict\_1": "0", \\
    "verdict\_2": "-1",\\
    ...\\
    \}\}\\
\}\}\\
\#\#\# \\ 
QUESTION:  Tell me about John.\\  
ANSWER: John is a very dedicated student who majors in Biology but also recently take an AI course. Other than being a student, he also works part-time.  \\
SOURCE:  \\
Document ID 1:\\ 
John is a student at XYZ University. He is pursuing a degree in Computer Science.\\ 

Document ID 2: \\
Image: A chart showing John is enrolled in several courses this semester, including Data Structures, Algorithms, and Database Management.\\

Document ID 3: \\
John is a diligent student and spends a significant amount of time studying and completing assignments. He often stays late in the library to work on his projects.\\  
RESULT:\\
\{\{\\
"answer\_statements":\\ 
    \{\{\\
    "statement\_1": "John is majoring in Biology.",\\
    "statement\_2": "John is taking a course on Artificial Intelligence.",\\
    "statement\_3": "John is a dedicated student.",\\
    "statement\_4": "John has a part-time job."\\
    \}\},\\
"verdicts":\\
    \{\{\\
    "verdict\_1": "0",\\
    "verdict\_2": "-1",\\ 
    "verdict\_3": "1",\\
    "verdict\_4": "-1"\\
    \}\}\\
\}\}\\
\#\#\#
}{Prompt for Multi-Modal LLM Consistency Metric (Part 1)}{llm_qa_multimodal_consistency_metric_pt_1}

\prompt{Prompt for Multi-Modal LLM Consistency Metric (Part 2)}{
\#\#\#\\
QUESTION: What is the Photosynthesis.  \\
ANSWER:  Albert Einstein was a genius.  \\
SOURCE:  \\
Document ID 1: \\
Photosynthesis is a process used by plants, algae, and certain bacteria to convert light energy into chemical energy.  \\
RESULT:  \\
\{\{\\
"answer\_statements":\\ 
    \{\{\\
    "statement\_1": "Albert Einstein was a genius."\\
    \}\},\\
"verdicts":\\
    \{\{\\
    "verdict\_1": "-1",\\
    \}\}\\
\}\}\\
\#\#\#\\  
QUESTION: What is the Photosynthesis.\\  
ANSWER:  Source talks about the Einstein theoretical physicist. The source does not provide any information on question.\\  
SOURCE: \\ 
Document ID 1:\\ 
Image: Page from a document containing a paragraph stating Albert Einstein was a German-born theoretical physicist who is widely held to be one of the greatest and most influential scientists of all time.\\
RESULT:  \\
\{\{\\
"answer\_statements":\\
    \{\{\\
    "statement\_1": "Source mentions Einstein is theoretical physicist",\\
    "statement\_2": "Source does not mention information on Photosynthesis."\\
    \}\},\\
"verdicts":\\
    \{\{\\
    "verdict\_1": "1",\\
    "verdict\_2": "1"\\
    \}\}\\
\}\}\\
\#\#\#  \\
"""
}{Prompt for Multi-Modal LLM Consistency Metric (Part 2)}{llm_qa_multimodal_consistency_metric_pt_2}

\end{document}